# AMPSO: Artificial Multi-Swarm Particle Swarm Optimization

Haohao Zhou[1], Zhi-Hui Zhan[2], Zhi-Xin Yang[3], Xiangzhi Wei[1,*]

[1] Institute of Intelligent Manufacturing and Information Engineering, Shanghai Jiao Tong University, Shanghai, 200240, China
[2] Guangdong Provincial Key Laboratory of Computational Intelligence and Cyberspace Information, School of Computer Science and Engineering, South China University of Technology, Guangzhou, 510006, PR China
[3] State Key Laboratory of Internet of Things for Smart City and Department of Electromechanical Engineering, University of Macau, Macao, China
*Corresponding author
Email: antonwei@sjtu.edu.cn

**Abstract:** In this paper we propose a novel artificial multi-swarm PSO which consists of an exploration swarm, an artificial exploitation swarm and an artificial convergence swarm. The exploration swarm is a set of equal-sized sub-swarms randomly distributed around the particles space, the exploitation swarm is artificially generated from a perturbation of the best particle of exploration swarm for a fixed period of iterations, and the convergence swarm is artificially generated from a Gaussian perturbation of the best particle in the exploitation swarm as it is stagnated. The exploration and exploitation operations are alternatively carried out until the evolution rate of the exploitation is smaller than a threshold or the maximum number of iterations is reached. An adaptive inertia weight strategy is applied to different swarms to guarantee their performances of exploration and exploitation. To guarantee the accuracy of the results, a novel diversity scheme based on the positions and fitness values of the particles is proposed to control the exploration, exploitation and convergence processes of the swarms. To mitigate the inefficiency issue due to the use of diversity, two swarm update techniques are proposed to get rid of lousy particles such that nice results can be achieved within a fixed number of iterations. The effectiveness of AMPSO is validated on all the functions in the CEC2015 test suite, by comparing with a set of comprehensive set of 16 algorithms, including the most recently well-performing PSO variants and some other non-PSO optimization algorithms.

Keywords: Swarm Intelligence, Particle Swarm Optimization, Multi-swarm

## 1. Introduction

The PSO algorithm was first proposed by Kennedy and Eberhart [1, 2] for solving optimization problems in 1995. This algorithm is motivated by the emergent motion of the foraging behavior of a flock of birds or a school of fish, where a member learns to move based on its own experience and the experiences of the others. PSO is simple in implementation and has nice convergence properties when compared to other evolutionary algorithms [3, 4], which makes PSO one of the most prominent population-based optimization techniques. Nowadays PSO has been successfully extended to many application areas such as communication networks [5, 6], medicine engineering [7], scheduling of tasks in cloud computing [8, 9], airport gate assignment [10], energy management [11], linguistics studies [12], and supply chain management [13].

Although PSO is considered as a robust algorithm with a fast convergence rate, it suffers from the premature convergence problem: PSO can be easily trapped into some local optima when solving multimodal problems [14, 15, 16, 17, 18, 19, 20, 21]. Therefore, a nice balance between exploration (global investigation of the search place) and exploitation (finer search around a local optimum) is critical to the success of PSO, especially for complex multimodal problems with a large number of local minima. In order to address this issue, a number of PSO variants have been developed. In the following we shall limit our survey to three major types: PSO with adjusted parameters, PSO with various topologies and PSO with hybrid strategies.

**PSO with Adjusted Parameters:** Adjusting $w$, $c_1$ and $c_2$ may also change the exploration and exploitation abilities of the PSO. For example, a larger $w$ facilitates the exploration while a smaller one encourages the exploitation. Zhan [14] proposed an adaptive PSO, which adjusted $w$, $c_1$ and $c_2$ based on an estimation of the evolutionary state that relied on the distribution of population and the fitness of particles. Zhang et al. [22] proposed a PSO variant using the Bayesian technique to adaptively change the inertia weight based on the previous positions of the particles such that the exploitation capability of PSO was enhanced. Tanweer et al. [23] proposed a PSO variant with the best particle conducting a self-regulation of its inertia weight to accelerate the searching towards the global optimum. Based on the fact that changing the inertia weight adaptively will change the stability conditions adaptively, Taherkhani et al. [24] proposed a PSO variant with stability-based adaptive inertia weight.

**PSO with Various Topologies:** It has been shown that different topologies of the particles significantly influenced the exploration and exploitation performances of PSO [25, 26, 27]. Kennedy [28] proposed a small-world social network and investigated the impact of different topologies of the particles (e.g., circles, wheels, stars and randomly-assigned edges) on the performances of PSO algorithms, and showed that sparse connected networks were suitable for addressing complex functions while dense connected network worked well for simple functions.

In 1999, Suganthan [29] first introduced the concept of dynamic topologies for PSO, where a particle was initially a subswarm by itself and the subswarms merged as the iterations went on, and eventually became a single fully-connected topology. Cooren et al. [27] proposed a particle swarm system called TRIBES, in which the topology evolved over time: the swarm was divided into sub-swarms each had its own structure, good tribes were allowed to get rid of their weakest members and bad tribes were allowed to add in a random number of new members to enhance their success rate.

Bonyadi et al. [30] proposed time-varying topologies by growing the sizes of the subswarms, merging the subswarms, and changing the topologies (e.g., global best topology, ring topology and non-overlap topology) of the subswarms as their sizes increased.

**PSO with Hybrid Strategies:** PSO has been combined with many other techniques for a better efficiency and accuracy, where the combined techniques are usually used to strengthen the local search of the computation process. For example, Mirjalili et al. [31] hybridized PSO with gravitational search algorithm for training feedforward neural networks. Zhang et al. [32] combined PSO with a back-propagation algorithm to train the weights of feedforward neural networks. Nagra et al. [33] developed a new hybrid of dynamic multi-swarm PSO with a gravitational search algorithm for improving the exploration and exploitation performance of PSO. Zhan et al. [34] combined PSO with orthogonal experimental design to discover the best combination of a particle's best historical position and its neighborhood's best historical position such that an excellent exemplar is constructed for the swarm to learn faster. Bonyadi et al. [30] hybridized PSO with a covariance matrix adaptation evolutionary strategy to improve the solutions at the latter phases of the evolution process. Garg [35] combined PSO with genetic algorithms (GA) such that the local search capability of GA can help speed up the convergence of PSO: for the best individual of the population after certain number of iterations, [35] created a new population by replacing weak points in the current population with excellent points via selection, crossover and mutation operators. Liu et al. [36] also combined PSO with GA to job scheduling issues. Plevris and Papadrakakis [37] combined PSO with a gradient-based quasi-Newton SQP algorithm for the optimization of engineering structures. N. Singh and S.B. Singh [38] combined PSO with Grey Wolf Optimizer for improving the convergence rate of the iterations. Raju et al. [39] combined PSO with a bacterial foraging optimization for FDM 3D printing parameters of complicated structures. Visalakshi and Sivanandam [40] combined PSO and the simulated annealing algorithm for processing dynamic task scheduling with load balance.

For interested readers, a comprehensive survey for hybrid PSO can be available in [41], and a recent survey of PSO with a broader range of techniques is available in [42].

**The major innovation of this study is as follows**: We proposed an artificial multi-swarm PSO (AMPSO) to improve the exploration and exploitation performances of PSO. The multi-swarm consists of an exploration swarm with a set of equal-sized sub-swarms, an exploitation swarm artificially generated around the neighborhood of the best particles produced by the exploration swarm for a fixed period of iterations, and finally a convergence swarm artificially generated around the neighborhood of the best particles produced by the exploitation swarm. Here by "artificial" it means that we generate a swarm as we need it instead of mimicking the natural evolutionary process of any social animal. A novel diversity scheme based on the positions and the fitness values of the swarms is proposed, and two update techniques are combined with the diversity scheme to guarantee nice results by using a fixed number of iterations.

We validate the effectiveness of AMPSO by comparing it with two recent PSO variants (as well as their corresponding sets of comparison algorithms) on the same set of CEC2015 benchmark functions.

The rest of the paper is presented as follows: Section 2 briefly reviews the PSO variants that are related to AMPSO; Section 3 presents our techniques of AMPSO;

Section 4 presents the comparison experiments with other PSO variants using the CEC2015 test suite; Section 5 concludes the paper with some discussions.

## 2. Related work

### 2.1. Standard PSO

The original PSO did not have an inertia weight [1, 2]. Since the inertia weight was introduced by Shi and Eberhart [43] in 1998, it has shown its power in better control of the exploration and exploitation processes.

$$v_i^d(t+1) = \omega v_i^d(t) + c_1 r_1 \left(p_i^d(t) - x_i^d(t)\right) + c_2 r_2 \left(p_g^d(t) - x_i^d(t)\right) \tag{1}$$

$$x_i^d(t+1) = x_i^d(t) + v_i^d(t+1) \tag{2}$$

When searching in a $D$-dimensional hyperspace, each particle $i$ has a velocity vector $v_i(t) = [v_i^1(t), v_i^2(t), \cdots, v_i^D(t)]$ and a position vector $x_i(t) = [x_i^1(t), x_i^2(t), \cdots, x_i^D(t)]$ for the $t$-th iteration. $p_i$ is the historically best position of the particle $i$, and $p_g$ is the global best particle of the swarm. Acceleration coefficients $c_1$ and $c_2$ are commonly set in the range [0.5, 2.5] (e.g., [44, 45]). $r_1$ and $r_2$ are two randomly generated values within range [0, 1], $\omega$ typically decreases linearly from 0.9 to 0.4 [43].

### 2.2. PSO with Multi-Swarms

Although competition among individual particles usually improves the computation performance, much greater improvements can be achieved through *cooperation* among multiple swarms. A large body of literature has been devoted to the cooperation among multiple warms. Due to the limit of space, in the following only a few typical multi-swarms PSO variants are reviewed.

Inspired by Potter's idea of splitting the solution vector into smaller vectors for improving the performance of the Genetic Algorithm [46], Bergh and Engelbrecht [47] proposed a cooperative approach for PSO by dividing a $d$-dimension swarm into $k$ ($k < d$) subswarms and exchanging the information (e.g., best particle) of each subswarm. Blackwell and Branke [48] proposed a multi-swarm PSO for benchmark functions where the peaks of the functions were dynamically changing. Liang's group [49, 50] proposed a dynamic multi-swarm system that used small-sized sub-swarms, which were randomly regrouped frequently by using various schedules and the information was exchanged among the subswarms; they also exploited the Quasi-Newton method to improve the local search performance of PSO. Xu et al. [3] proposed a dynamic multi-swarm PSO with cooperative learning strategy, where each dimension of the two worst particles learned from two better sub-swarms using tournament selection. Chen et al. [20] proposed a dynamic multi-swarm PSO with a differential learning strategy, where the exploitation capability of PSO was enhanced by combining the differential mutation into PSO and employing Quasi-Newton method as a local searcher. Zhang et al. [51] proposed a self-adaptive and cooperative PSO based on four sub-swarms. The cooperative capability was guaranteed by sharing the unique global best. Zhao [52] proposed a dynamic multi-swarm particle swarm optimizer (DMS-PSO) with harmony search (HS), and merged HS into each sub-swarm to make use of exploration capability of the DMS-PSO and stochastic exploitation capability of HS. Ye et al. [53] proposed a novel multi-swarm PSO with two dynamic learning strategies of updating the velocities of the particles, of which one was turning normal particles into communication particles as the iteration number increases and the other was updating the velocities based on the local best of each subswarm; the latter was inspired by the full-informed mechanism of [26] that took the information of all the neighbors (instead of the global best particle) into consideration. Recently, Wang et al. [54] proposed a multi-swarm distributed PSO whose size of sub-swarm was adaptively control by an adaptive granularity learning strategy.

## 3. Artificial Multi-Swarm Particle Swarm Optimization (AMPSO)

In this section, the technical details of AMPSO are presented in the following manner: the main idea of AMPSO is presented in subsection 3.1, the diversity and parameter adjustment techniques are presented in subsection 3.2, the techniques for the generation of artificial swarms are presented in subsection 3.3, and finally the swarm reconstruction techniques are presented in subsection 3.4.

### 3.1. Main Idea of AMPSO

Our idea of artificial multi-swarm PSO is inspired by the following rule of the game for capturing the commander of the enemy that is widely used in modern wars:

*First of all, attackers send out many small searching teams to locate the suspected regions that the commander may stay in; and then a fast-moving air force is sent to attack these regions and determines highly suspected regions; and finally, a ground force with strong fire is sent to sweep each highly suspected region to capture the commander of the enemy.*

Based on this idea, we name our three swarms as *Exploration Swarm*, *Exploitation Swarm* and *Convergence Swarm*. This naming is consistent with the three evolutionary states proposed in APSO [14]: *exploration state*, *exploitation state* and *convergence state*. Here the exploration swarm consists of a set of randomly distributed small-sized sub-swarms searching through the space, and each sub-swarm evolves independently (without any communication among sub-swarms) in order to explore as much searching space as possible; the exploitation swarm is generated from a random perturbation of the position of the best particle of the exploration swarm after a number of iterations. It is possible that the selected "best particle" is in reality a locally best particle rather than the globally best particle. To increase of chance of achieving the global best, when the exploitation swarm runs into a stagnation state the exploration swarm is reinitialized and a sequence of iterations will output a new best particle, which is then perturbed into a new exploitation swarm; meanwhile, when the exploitation swarm is in a stagnation state its best particle is perturbed into the convergence swarm. Figure 1 illustrates the procedure.

Although we exploit the concept of APSO [14] for different evolutionary states, our approach is quite different in the following aspects: instead of using one swarm, we use multiple swarms; and our swarms are artificially generated around hopeful regions to sweep the targets rapidly. In the experiment section, we shall show that our strategy is fairly promising in achieving better results when compared with the existing popular PSO variants.

For ease of description, we present the definition of the parameters used in the rest of the paper in Table 1.

**Table 1**
Nomenclature

| | |
|---|---|
| $N_{er}, N_{ei}, N_c$ | the size of exploration swarm, the size of exploitation swarm and the size of convergence swarm, respectively |
| $N_{total}, N_i, N_1, N_2$ | the total number of iterations, the total number of iterations consumed so far, the maximum number of iterations for the Exploration Swarms, and the maximum number of iterations for the Exploitation Swarm, respectively |
| $N_s$ | the number of weak particles of the exploitation swarm to be updated |
| $t$ | the current index of the iteration |
| $D$ | the dimension of the solution space |
| $v_i(t)$ | the velocity of particle $i$ in the $t$-th iteration |
| $x_i(t)$ | the position of particle $i$ in the $t$-th iteration |
| $\omega, c_1, c_2$ | initial weight and accelerations for PSO, respectively |
| $r_1, r_2$ | two random real numbers in range [0, 1] |
| $Er(t)$ | the evolution rate of the swarm in the $t$-th iteration |
| $bf(t)$ | the globally best fitness of the swarm in the $t$-th iteration |
| $K$ | a user-defined constant used for computing $Er(t)$ |
| $\alpha_1, \alpha_2, \alpha_3$ | ratios used to control $N_1$, $N_2$ and $N_s$ respectively and $\alpha_1, \alpha_2, \alpha_3 \in [0,1]$ |
| $\beta$ | a parameter used to control the termination condition of the exploitation swarm: the condition is satisfied if $Er(t) < \beta$ |
| $E$ | the entropy that represents the swarm's diversity degree |
| $E_d(t), E_p(t), E_f(t)$ | the diversity of the $d$-th dimension of the swarm, the position diversity of the swarm, and the fitness diversity of the swarm in the $t$-th iteration, respectively |
| $dr$ | a $D$-dimension vector $dr[dr^1, dr^2, ..., dr^D]$, where $dr^i$ is a random real number and $dr^i \in N(0, (0.1)^2)$ |
| $N_{ne}$ | the number of non-progress iterations |
| $P_c$ | the probability of reconstructing the Convergence Swarm |
| $x_{lb}, x_{ub}$ | the lower bound and upper bound of the search space respectively |

Usually the evolution rate is calculated as the difference between two consecutive iterations. However, due to the slow evolution of the particles between two consecutive iterations it is possible that the evolution rate is fairly small and cannot be used to measure the progress of the particles. To mitigate this, refer to Eq. (3), we take the evaluation rate as the average evolution value of $K$ iterations divided by the evaluation of the previous iteration. To guarantee that $K$ is chosen reasonably such that the evaluation rate is large enough while the frequency of the evaluations is large enough, assume without loss of generality that the maximum number of iterations (i.e., $N_{er}$, or $N_{ei}$, or $N_c$) is larger than 1000, we set $K=50$. In general, $K$ can be determined by a simple experiment with some test functions. Formally, we have the expression of the evaluation rate in Eq. (3).

$$Er(t) = \begin{cases} \dfrac{bf(t-K) - bf(t)}{K * bf(t-1)} & t \geq K \\ \dfrac{bf(1) - bf(t)}{t * bf(t-1)} & 1 < t < K \\ 1 & t = 1 \end{cases} \quad (3)$$

Refer to Figure 1, AMPSO works as follows:

The AMPSO consists of three modules, the Exploration Swarm, the Exploitation Swarm and the Convergence Swarm. After an initialization of the Exploration Swarm (details are presented in Section 4), it evolves according to Eq. (1-2). As the total number of iterations $N_i$ surpasses $N_{total}/3$, the Convergence Swarm starts to work. Before triggering the Convergence Swarm, the Exploration Swarm and the Exploitation Swarm alternatively evolve and search the solution space for all possible local optima.

In this process, the Exploration Swarm consists of many equal-sized small sub-swarms searching around the solution space and provides with the Exploitation Swarm small regions around each locally best candidate to focus on; and each sub-swarm iterates $N_1$ times, evaluate diversity $E$ and updates inertia weight $\omega$ adaptively. Since the Exploration Swarm requires a stronger capability of searching around the solution space, a large diversity is maintained for it. After the Exploration Swarm runs out of $N_1$ iterations, its best particle is perturbed into an Exploitation Swarm. To save the effort of function evaluations (FEs) and guarantee the computation efficiency, the evolution of the Exploitation Swarm is forced to stop as the number of iterations exceeds the maximum allowed quota or the evolution rate is smaller than a threshold, i.e., $t > N_2 (N_2 = \alpha_2 N_{total})$ or $Er(t) < \beta$.

During the exploitation process of the Exploitation Swarm, a small diversity is required to make the swarm rapidly converge. To guarantee this, at each iteration $N_s$ weak particles are updated using the perturbation of the best particle's position.

After this operation, $N_s$ operations has been consumed. To maintain the same number of operations for each iteration and therefore a fixed number of iterations for AMPSO, out of the new swarm (of size $N_{ei}$) only $(N_{ei} - N_s)$ particles are selected randomly to be updated by Eq. (1-2). The exploration and exploitation processes alternate as $N_i > (N_{total}/3)$.

Similar to the generation of the Exploitation Swarm, the Convergence Swarm is also generated from a perturbation of the best particle's position. Using a fixed total of $N_{total}$ iterations, the remaining number of iterations for the Convergence Swarm is thus $(N_{total} - N_i)$. To guarantee a better result, different from the Exploitation Swarm which emphasizes on rapid convergence, the convergence process balances the exploration and the exploitation states by using a moderate diversity with a new setting of $\omega$. In addition, to help the Convergence Swarm jump out of the trap of local optima, a swarm reconstruction strategy is invoked with a probability $P_c$. The "Reconstruct the swarm" and "Update the swarm" in the convergence module of the flowchart (Figure 1) are conducted as follows: Normally, the swarm is "updated" according to Eq. (1-2); when the swarm is trapped in a locally optimal region (evaluated by $P_c$), to jump out of the trap all particles in the swarm are reconstructed by using the best particle's position, the details of the reconstruction technique shall be presented in Section 3.4.

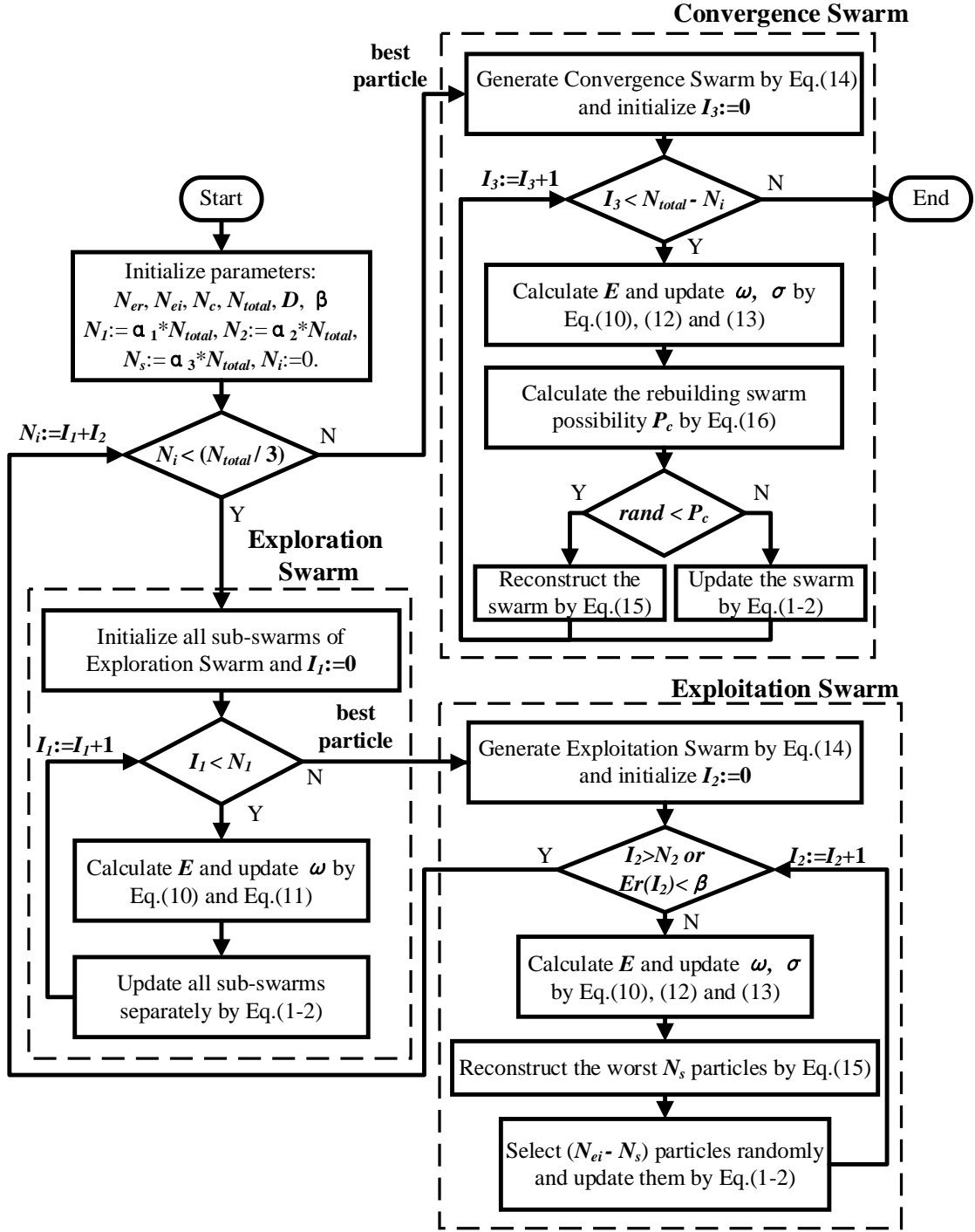

Fig. 1. AMPSO flowchart

## 3.2. Diversity and Parameter Adjustment

Diversity has been known as an especially critical factor for controlling the exploration and exploitation abilities of the swarm [16, 55, 56, 57, 58]. Lack of population diversity is an important factor for premature convergence of the swarm at a local optimum [2]. And enhancing diversity was considered to be a useful approach to escape from the local optima [2, 14]. However, enhancing the swarm diversity may also require a larger number of iterations to achieve the global best solution. In the following sections, we shall introduce our novel strategy of diversity measurement and the parameter settings for AMPSO in details, which achieves closer solutions to the global optima of a large number of complex functions in CEC2015 test suite by using the same amount of iterations compared with existing peer algorithms.

### 3.2.1. A Novel Diversity Strategy

Diversity measures the distribution state of the particles: when the particles are gathering near some local optima (which has the potential to be global best), the difference (e.g., positions and fitness values) and therefore the diversity among the particles are small; on the other hand, when the particles are roaming around the searching space, the diversity is large. Therefore, we can control the exploration and exploitation states of a swarm by adjusting the parameters that affect the diversity of the swarm.

To obtain nice diversities, two major metrics have been proposed: the population distribution entropy (denoted as PDE) [59, 60, 61] and average distance amongst points (denoted as ADAP) [62].

Of the two types of methods ADAP is rather straightforward, and it can be further categorized into two methods: one method is based on the average distance from the swarm center [63] (Eq. (4)) and the other method is based on the average of the average distance in between each pair of particles in the swarm [64] (Eq. (5)).

$$diversity(t) = \frac{1}{N*L} \sum_{i=1}^{N} \sqrt{\sum_{k=1}^{D}(x_i^k - \overline{x^k})^2} \qquad (4)$$

$$diversity(t) = \frac{1}{N*L} \sum_{i=1}^{N} \left( \frac{1}{N} \sum_{j=1}^{N} \sqrt{\sum_{k=1}^{D}(x_i^k - x_j^k)^2} \right) \qquad (5)$$

where $N$ is the number of poarticles in the swarm, and $L$ is the length of the longest diagonal in the search space.

In addition, the scheme of population-distribution-entropy was originally motivated by the principle of Shannon entropy in information theory [65]. The basic concept of this method is as follows: the current search space is divided into $Q$ parts with equal size; let $Z_i$ be the number of particles in the $i$-th part, let $p_i = Z_i/N$ be the probability that the particles are contained in the $i$-th area. Then, taking population-distribution-entropy as our diversity, the population diversity can be defined by Eq. (6-7).

$$p_i = \frac{Z_i}{N} \qquad (6)$$

$$E = -\sum_{i=1}^{Q} p_i \log_n p_i \tag{7}$$

Based on ADAP and PDE a dozen of algorithms have been proposed [66, 67, 68]. Particularly, Zhan [14] proposed an Evolutionary Factor computed by comprehensively considering the distance between each pair of particles and the distance from the best particle to each of the remaining particles.

However, the ADAP and PDE methods are both based on the distance metric, which may lead to misguiding entropy information. For example, for a multimodal function with multiple local optima ADAP might yield a very large entropy when the particles are near a number of local optima, but a small entropy is needed in this case for the swarm to converge rapidly to the global best. PDE may lose the distribution information of the positions and fitness values of the particles. For example, the mutual Euclidean distance metric three points [1,0,0], [0,1,0] and [0,0,1] is $2^{0.5}$, and their Euclidean distance entropy (i.e., the degree of the difference of Euclidean distances between every pair of points) is 0, which cannot reflect the position distribution and fitness distribution of the particles.

To avoid the above-mentioned problems due to the distance metric, in this paper we propose a novel hybrid of the position diversity and the fitness diversity as follows:

**Position Diversity:** First of all, we divide each position dimension $d$ into $Q$ parts, count the number of particles that fall into the $i$-th part and compute $p_{id}$ by Eq. (6), and finally the position diversity of the $d$-th dimension and the position diversity of the entre swarm can be obtained by Eq. (8) and Eq. (9) respectively.

$$E_d(t) = -\sum_{i=1}^{Q} p_{id} \log_n p_{id} \tag{8}$$

$$E_p(t) = \frac{1}{D} \sum_{d=1}^{D} E_d(t) \tag{9}$$

**Fitness Diversity:** Similar to the position diversity, the distribution of the fitness values of the particles, i.e., the fitness diversity of the swarm, also reflects the exploration or convergence state of the swarm. Particularly, when a swarm gathers around some local optima (which has the potential to be global best) the fitness diversity is fairly small; on the other hand, when the swarm is distributed randomly it can roam around the searching space and the fitness diversity is relatively large.

For example, in Figure 2, although the positions of the four points in Figure 2(a) and Figure 2(b) are the same, their fitness values vary significantly. Therefore, to guide the search towards the global best, we take a combination of the position diversity and fitness diversity into consideration.

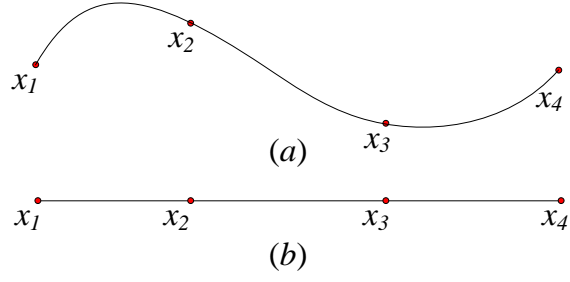

Fig. 2. The difference between position diversity and fitness diversity:
(a) points with different fitness, (b) points with the same fitness.

Our calculation of the fitness diversity of a swarm is similar to that of [61]. More precisely, we divided $[fitness_{min}, fitness_{max}]$ into $Q$ parts and calculate $p_i$ for part $I$, and then obtain $E_f(t)$ by Eq. (7). Formally, our hybrid of position diversity and fitness diversity is formulated in Eq. (10).

$$E(t) = \frac{E_p(t) + E_f(t)}{2} \tag{10}$$

**3.2.2. Parameter Adjustment**

The inertia weight plays a critical role in balancing the exploration and exploitation capabilities of the swarm, and it also significantly affects the accuracy of the results [22, 23, 24]. Therefore, we shall adaptively adjust the inertia weight function to guarantee the effective cooperation of the swarms. Meanwhile, for ease of computation, we fix $c_1 = c_2 = 1.49445$ based on a large set of experimental results of previous PSO variants (e.g., [22, 50, 53]).

Since the Exploration Swarm, the Exploitation Swarm and the Convergence Swarm have different functions during the searching process for the global best, different control parameters will be set for the swarms. Specifically, the detailed strategies for the swarms are as follows:

**(1) Exploration Swarm**

The role of the swarm is to explore the searching space broadly and provide small searching regions for the Exploitation Swarm. For this purpose, the Exploration Swarm is divided into a set of equal-sized sub-swarms randomly distributed around the solution space (i.e., they are not necessarily neighboring to each other) and each sub-swarm evolves independently (without communication) in order to explore the searching space as much as possible. For the inertia weight $w$, it non-linearly decreases from 0.9 to 0.6 to maintain a large diversity $E(t)$ (with a strong exploration ability). Formally, using a sigmoid-like mapping we have Eq. (11) for the Exploration Swarm.

$$\omega(t) = \frac{1}{1+0.67e^{-2.67E(t)}} \in [0.6, 0.9] \quad \forall E(t) \in [0,1] \tag{11}$$

### (2) Exploitation Swarm

The role of the swarm is rapidly converging to the neighborhood of the local best solutions without emphasizing too much on the precision level. We also design a sigmoid-like update mechanism to change the inertia weight adaptively from 0.8 to 0.5 in order to improve the exploitation ability of Exploitation Swarm. In addition, to further accelerate the convergence process, a partial swarm reconstruction strategy is adopted by replacing the positions of the worst $N_s$ particles from a perturbation of the best particle's position. To control the range of distribution, a Gaussian distribution $N(0, \sigma^2)$ is exploited, where $\sigma$ decreases from 0.2 to 0.1. Formally we have Eq. (12-13) for controlling the Exploitation Swarm.

$$\omega(t) = \frac{1}{1 + e^{-ln4*E(t)}} \in [0.5, 0.8] \quad \forall E(t) \in [0,1] \quad (12)$$

$$\sigma(t) = \frac{1}{1 + 9e^{(ln4-ln9)*E(t)}} \in [0.1, 0.2] \quad \forall E(t) \in [0,1] \quad (13)$$

### (3) Convergence Swarm

In this scenery, the role of the swarm is converging to a higher accuracy solution at a proper speed and jumping out of the trapes of local optimal regions. In order to obtain a nice solution, a large inertia weight corresponding to the strong exploration ability of the swarm is needed at the beginning and it should decrease to a small value that corresponds to the ability of refining solution as the diversity decreases, therefore we set the range of inertia weight as [0.5, 0.8] by Eq. (12). As for jumping out of local traps, the Convergence Swarm utilizes a swarm reconstruction strategy. Different from the partial swarm reconstruction strategy for Exploitation Swarm that replaces $N_s$ worst particles, the positions of all particles are replaced from the perturbation of the best particle. This helps the swarm jump out from a local trap rapidly. Similarly, a Gaussian distribution $N(0, \sigma^2)$ is exploited to control the range of distribution, where $\sigma$ is also determined by Eq. (13).

### 3.3. Generation of Artificial Swarms

In our strategy, the Exploitation Swarm is an artificial swarm generated from the best particle of the Exploration Swarm, and the Convergence Swarm is an artificial swarm generated from the best particle of the Exploitation Swarm. Since the chosen best particle is near some local optimal point, to increase the chance of achieving the local best, we shall perturb the position of the best particle $x[x^1, x^2, \ldots, x^D]$ into a set of particles around a small neighborhood of $x$ by Eq. (14).

$$x' = x + (x_{ub} - x_{lb})dr \quad (14)$$

where $dr = [dr^1, dr^2, \ldots, dr^D]$, and $dr^i$ follows the Gaussion distribution with $dr^i \in N(0, (0.1)^2)$, and $x \in [x_{lb}, x_{ub}]$.

### 3.4. Swarm Reconstruction

**Partial Swarm Reconstruction for Exploitation:** To guarantee that the Exploitation Swarm rapidly converges to the neighborhood of a promising local best that has a potential to become the global best, apart from the adjustment of the parameters, in each iteration we also force the lousiest $N_s$ to change their positions according to the best particle's position. Different from the technique of artificial swarm generation by perturbing *dr* for each dimension, the reconstruction technique here alters merely one dimension randomly. This is based on the consideration that the best particle wins out because quite a few of its dimensions have reached very nice states and changing merely one dimension may preserve its nice property as much as possible. As mentioned in Section 3.1, to guarantee a constant number of update operations in each iteration and therefore a fixed number of total iterations, only $(N_{ei} - N_s)$ random particles are updated. Formally, the reconstruction technique alters merely one random dimension of the position vector works as follows:

$$x_{new}^d = x^d + (x_{ub} - x_{lb})\, r \tag{15}$$

where $d$ is a randomly chosen dimension, $r$ follows the Gaussian distribution and $r \in N(0, \sigma^2)$. $\sigma$ can be updated by Eq. (13) to yield nice results.

**Swarm Reconstruction for Convergence:** For the Convergence Swarm, when the evolution rate of the swarm gets stagnated for a consecutive $N_{ne}$ iteration, the swarm needs to jump out of a local trap. Using the best particle's position, we reconstruct the entire swarm based on a probability $p_c$ as follows:

$$P_c = \frac{1}{1 + e^{(0.01 * N_{total} - N_{ne})}} \tag{16}$$

where $N_{ne}$ is increased by one for every new iteration when $Er(t) < \beta$.

### 4. Experiments and Comparisons

To validate our AMPSO on complex multimodal functions, we choose CEC2015 benchmark functions [69] as our test suite. Refer to Table 2, this test suite consists of the latest collection of 15 unconstrained continuous optimization problems with various difficulty levels. According to their properties, these functions are divided into four groups: unimodal functions, simple multimodal functions, hybrid functions, and composition functions. All the guidelines of CEC2015 have been strictly followed for the experiments. Following the instructions of the test suite, every test of a function is conducted for 30 runs independently. The results for the 10-*D* case and the 30-*D* case are calculated. The range of the search space for each dimension is [−100, 100]. Population initialization is generated uniformly in the specified search space by a random number generator with clock time. The iteration process terminates as the maximum function evaluation FEs = 10000*D* is reached. The fitness value is *Fi*(*x*) − *Fi*(*x\**) after the maximum iteration is reached, where *Fi*(*x\**) is just a number about the corresponding function for instruction.

Table 2
Summary of the CEC2015 benchmark functions

| Type | No. | Function name | $F_i^*=F_i(x^*)$ |
|---|---|---|---|
| Unimodal Functions | 1 | Rotated High Conditioned Elliptic Function | 100 |
| | 2 | Rotated Cigar Function | 200 |
| Simple Multimodal Functions | 3 | Shifted and Rotated Ackley's Function | 300 |
| | 4 | Shifted and Rotated Rastrigin's Function | 400 |
| | 5 | Shifted and Rotated Schwefel's Function | 500 |
| Hybrid Functions | 6 | Hybrid Function ($N = 3$) | 600 |
| | 7 | Hybrid Function ($N = 4$) | 700 |
| | 8 | Hybrid Function ($N = 5$) | 800 |
| Composition Functions | 9 | Composition Function 1 ($N = 3$) | 900 |
| | 10 | Composition Function 2 ($N = 3$) | 1000 |
| | 11 | Composition Function 3 ($N = 5$) | 1100 |
| | 12 | Composition Function 4 ($N = 5$) | 1200 |
| | 13 | Composition Function 5 ($N = 5$) | 1300 |
| | 14 | Composition Function 6 ($N = 7$) | 1400 |
| | 15 | Composition Function 7 ($N = 10$) | 1500 |

To validate the effectiveness of our algorithm, using CEC2015 as their test suite, a set of comparison experiments with 4 recently published PSO variants and the sets of comparison algorithms that they used for comparisons purposes. The information of these algorithms is provided in Table 3.

Table 3
The Algorithms for Comparison

| | | |
|---|---|---|
| 1 | A modified particle swarm optimizer | GPSO [43] |
| 2 | Population structure and particle swarm performance | LPSO [25] |
| 3 | Defining a standard for particle swarm optimization | SPSO [70] |
| 4 | Comprehensive learning particle swarm optimizer | CLPSO [71] |
| 5 | The fully informed particle swarm | FIPS [26] |
| 6 | Dynamic multi-swarm particle swarm optimizer with local search | DMSPSO [49] |
| 7 | Modified particle swarm optimization and its applications | MPSO [72] |
| 8 | Traditional global PSO algorithms with inertia weight | GPSO*[1] |
| 9 | PSO with multiple subpopulations | MSPSO [73] |
| 10 | The fully informed particle swarm | FIPS [26] |
| 11 | PSO with dynamic tournament topology strategy | DTTPSO [74] |
| 12 | Dynamic multi-swarm particle swarm optimizer with local search | DMSPSO [49] |
| 14 | Artificial Bee Colony | ABC [75] |
| 15 | Teaching-Learning-Based Optimization | TLBO [76] |

| 16 | Multi-swarm PSO with dynamic learning strategy | PSODLS [53] |
| 17 | Ensemble particle swarm optimizer | EPSO [77] |
| 18 | Heterogeneous comprehensive learning PSO | HCLPSO [78] |

Note: GPSO* on line 8 and GPSO on line 1 are two distinct PSO variants cited from [1] and [43] respectively. FIPS on line 5 and 10 and DMSPSO on line 6 and 14 are the same but are tested under slightly different criteria

First of all, using the same comparison criterial with MPSO [72], we compare our AMPSO with MPSO and its corresponding comparison algorithms (GPSO, LPSO, SPSO, CLPSO, FIPS, DMSPSO). The same 12 representative functions ($f_1$-$f_9$, $f_{11}$, $f_{14}$ and $f_{15}$) that consist of 2 unimodal functions ($f_1$ and $f_2$), 3 multimodal functions ($f_3$, $f_4$ and $f_5$), 3 hybrid functions ($f_6$, $f_7$ and $f_8$) and 4 composition functions ($f_9$, $f_{11}$, $f_{14}$ and $f_{15}$) from CEC2015 test suite [69] are used tested, and the performance of the algorithms is characterized by statistical means and standard deviations, which is exactly the same criteria as used in [72]. For all the algorithms, the swarm size is set to 40. So, the maximum number of iterations is 2500 for 10$D$ and 7500 for 30$D$. The statistical results for the test functions in 10-$D$ and 30-$D$ are summarized in Table 4 and Table 5 respectively. Note that the lowest mean and standard deviation in each line are the best results and are highlighted in bold.

Secondly, using the same criteria with PSODLS [53], we compare our AMPSO with PSODLS and its corresponding comparison algorithms (GPSO*, MSPSO, FIPS, DTTPSO, DMSPSO, ABC, TLBO). The same 15 functions from CEC2015 all are tested and the same statistical criteria (i.e., best, worst, average, and median) are evaluated. For all the algorithms, the swarm size is set to 40. The maximum number of iterations is 2500 for 10$D$ and 7500 for 30$D$. Note that this is also the same setting in [72]. Table 6-9 summarize the statistical results.

Finally, we compare our AMPSO with two recent PSO variants, EPSO [77] and HCLPSO [78]. For completeness, we use the statistics of mean, standard deviation, best, worst, and median. Table 10 summarizes the statistical results.

To conduct the comparison experiments, some control parameters of our AMPSO are set as follows: $N_{er} = 10$, $N_{ei} = N_c = 40$ (same as the setting of MPSO [72] and PSODLS [53]), $N_{total}$ = FEs / $N_c$, $\alpha_1 = 0.02$, $\alpha_2 = 0.2$, $\alpha_3 = 0.25$, $\beta = 0.001$, $v_{max} = 0.01 * (x_{ub} - x_{lb})$.

**Table 4**
Comparison of AMPSO with MPSO and its comparison suite for 10-*D* case

| Func | Item | GPSO | LPSO | SPSO | CLPSO | FIPS | DMSPSO | MPSO | AMPSO |
|---|---|---|---|---|---|---|---|---|---|
| F1 | Mean | 5.05E+06 | 1.17E+05 | 2.78E+04 | 5.53E+05 | 2.51E+05 | 9.64E+04 | 2.63E+04 | **3.59E+03** |
|    | Std  | 2.31E+07 | 7.30E+04 | 1.01E+04 | 7.42E+04 | 3.05E+04 | 2.46E+05 | 6.61E+03 | **5.01E+03** |
| F2 | Mean | 6.02E+08 | 2.94E+07 | 6.15E+03 | 4.73E+04 | 6.49E+03 | 1.38E+04 | 6.36E+03 | **2.76E+02** |
|    | Std  | 1.05E+09 | 7.79E+07 | 2.83E+04 | 8.12E+04 | 3.86E+02 | 5.47E+03 | **2.12E+02** | 3.49E+02 |
| F3 | Mean | 2.03E+01 | 2.02E+01 | 2.02E+01 | 2.02E+01 | 2.03E+01 | 2.01E+01 | **1.87E+01** | 1.94E+01 |
|    | Std  | 1.12E-01 | 4.07E-02 | 2.99E-02 | 2.41E+00 | 1.51E-01 | 1.46E-01 | **1.18E-02** | 3.14E+00 |
| F4 | Mean | 1.48E+01 | 1.35E+01 | 4.87E+00 | 1.02E+01 | 6.75E+00 | 9.19E+00 | **3.23E+00** | 1.41E+01 |
|    | Std  | 1.12E+01 | 9.61E+00 | 4.89E+00 | 3.73E-01 | 1.72E+00 | 6.16E+00 | **1.76E-02** | 4.26E+00 |
| F5 | Mean | 5.76E+02 | 2.91E+02 | 4.14E+02 | 6.11E+02 | 5.14E+02 | 3.34E+02 | **1.89E+02** | 2.62E+02 |
|    | Std  | 9.36E+01 | 8.95E+02 | 6.49E+02 | 1.46E+02 | 4.58E+02 | 1.76E+02 | **8.85E+01** | 1.54E+02 |
| F6 | Mean | 4.69E+03 | 5.83E+03 | 1.13E+03 | 1.72E+03 | 7.11E+02 | 1.78E+03 | 5.20E+02 | **2.37E+02** |
|    | Std  | 6.41E+03 | 1.04E+04 | 5.75E+03 | 3.96E+02 | 7.15E+02 | 1.38E+04 | 3.17E+02 | **1.88E+02** |
| F7 | Mean | 5.56E+00 | 3.06E+00 | 1.34E+00 | 1.53E+00 | 8.87E-01 | 1.96E+00 | 8.99E-01 | **6.22E-01** |
|    | Std  | 2.12E+00 | 4.30E-01 | 1.99E-01 | 4.24E-01 | 2.39E-01 | 4.07E-01 | **1.23E-01** | 3.69E-01 |
| F8 | Mean | 6.53E+03 | 1.81E+03 | 1.27E+03 | 4.87E+02 | 6.89E+02 | **1.79E+02** | 4.46E+02 | 2.39E+02 |
|    | Std  | 4.98E+04 | 3.51E+03 | 2.54E+03 | 2.88E+02 | 1.67E+03 | **6.33E+01** | 1.28E+02 | 3.35E+02 |
| F9 | Mean | 1.07E+02 | 1.01E+02 | **1.00E+02** | **1.00E+02** | **1.00E+02** | **1.00E+02** | **1.00E+02** | **1.00E+02** |
|    | Std  | 1.79E+01 | 1.01E+00 | 7.99E-02 | 1.39E-02 | 1.92E-01 | 5.41E-01 | **1.06E-02** | 8.40E-02 |
| F11| Mean | 3.14E+02 | 2.59E+02 | 2.15E+02 | **1.62E+01** | 4.64E+01 | 1.31E+02 | 1.90E+02 | 1.26E+02 |
|    | Std  | 2.22E+01 | 1.11E+02 | 5.62E+02 | **5.15E+00** | 1.13E+02 | 4.99E+02 | 2.66E+01 | 1.45E+02 |
| F14| Mean | 6.25E+03 | 5.47E+03 | 6.19E+03 | 2.45E+03 | 2.29E+03 | 3.17E+03 | 2.71E+03 | **9.34E+01** |
|    | Std  | 4.04E+06 | 6.32E+03 | 8.62E+03 | 4.51E+02 | 1.70E+03 | 2.89E+03 | 7.25E+02 | **2.53E+01** |
| F15| Mean | 1.25E+02 | 1.06E+02 | **1.00E+02** | **1.00E+02** | **1.00E+02** | **1.00E+02** | **1.00E+02** | **1.00E+02** |
|    | Std  | 8.43E+00 | 4.05E+01 | **0.00E+00** | **0.00E+00** | 1.73E-12 | 2.63E-13 | 5.68E-19 | 1.79E-09 |

**Table 5**
Comparison of AMPSO with MPSO and its comparison suite for 30-*D* case

| Func | Item | GPSO | LPSO | SPSO | CLPSO | FIPS | DMSPSO | MPSO | AMPSO |
|---|---|---|---|---|---|---|---|---|---|
| F1 | Mean | 2.72E+08 | 3.52E+07 | 2.23E+05 | 5.56E+06 | 4.00E+06 | 6.72E+06 | 1.79E+06 | **4.41E+04** |
|  | Std | 3.54E+08 | 5.92E+07 | 2.42E+05 | 5.80E+06 | 2.42E+05 | 1.15E+07 | 2.38E+05 | **2.08E+04** |
| F2 | Mean | 2.23E+10 | 1.91E+09 | 3.70E+03 | 4.48E+03 | 6.29E+03 | 3.37E+03 | 1.46E+03 | **6.60E+02** |
|  | Std | 1.96E+09 | 2.09E+09 | 2.44E+04 | 1.70E+03 | 1.43E+04 | 2.92E+03 | **7.28E+01** | 1.45E+03 |
| F3 | Mean | 2.08E+01 | 2.08E+01 | 2.09E+01 | 2.09E+01 | 2.10E+01 | 2.05E+01 | 2.03E+01 | **2.00E+01** |
|  | Std | 7.54E-03 | 3.58E-01 | 8.30E-02 | 1.90E-02 | 9.90E-02 | 1.53E-01 | **2.33E-03** | 8.68E-03 |
| F4 | Mean | 1.55E+02 | 1.03E+02 | 3.55E+01 | 9.02E+01 | 1.54E+02 | 8.23E+01 | **1.91E+01** | 1.47E+02 |
|  | Std | 8.14E+01 | 1.29E+01 | 3.38E+00 | 3.63E+00 | 4.96E+01 | 1.54E+01 | **4.06E-01** | 2.24E+01 |
| F5 | Mean | 3.54E+03 | 3.02E+06 | 3.97E+03 | 4.62E+03 | 6.31E+03 | 3.79E+03 | 2.62E+03 | **2.31E+03** |
|  | Std | 3.22E+02 | 8.32E+02 | 1.39E+03 | 6.03E+02 | 1.51E+03 | 5.40E+02 | **1.24E+02** | 3.55E+02 |
| F6 | Mean | 1.00E+07 | 1.30E+06 | 1.14E+05 | 3.53E+05 | 4.37E+05 | 1.70E+05 | 4.41E+04 | **2.20E+04** |
|  | Std | 7.16E+07 | 5.56E+05 | 3.13E+04 | 1.71E+05 | 5.05E+05 | 8.56E+04 | **7.17E+03** | 2.25E+04 |
| F7 | Mean | 4.77E+01 | 2.46E+00 | 9.20E+00 | 9.10E+00 | 1.24E+01 | 1.29E+01 | 9.98E+00 | **7.76E+00** |
|  | Std | 3.55E+01 | 3.69E+01 | 1.25E+00 | 8.60E-01 | 4.09E+00 | **9.84E-02** | 3.15E+00 | 9.65E-01 |
| F8 | Mean | 1.73E+06 | 2.36E+05 | 3.22E+04 | 6.36E+04 | 4.73E+04 | 7.95E+04 | 2.64E+04 | **1.32E+04** |
|  | Std | 2.05E+07 | 4.11E+05 | 2.60E+04 | 9.24E+04 | 9.13E+03 | 1.91E+05 | **3.36E+03** | 6.14E+03 |
| F9 | Mean | 2.15E+02 | 1.29E+02 | 1.03E+02 | 1.04E+02 | 1.03E+02 | 1.04E+02 | **1.03E+02** | 1.04E+02 |
|  | Std | 8.37E+01 | 5.08E+00 | 1.98E-01 | 1.81E-01 | 8.50E-02 | 7.00E-02 | **8.07E-03** | 3.82E-01 |
| F11 | Mean | 1.24E+03 | 1.04E+03 | 5.91E+02 | 3.55E+02 | 4.39E+02 | 5.98E+02 | 6.03E+02 | **3.04E+02** |
|  | Std | 3.88E+02 | 2.09E+02 | 7.22E+01 | 4.81E+01 | 8.05E+01 | 7.56E+02 | 1.88E+01 | **1.29E+00** |
| F14 | Mean | 4.86E+04 | 3.85E+04 | 3.36E+04 | 2.89E+04 | 2.73E+04 | 3.07E+04 | 2.73E+04 | **7.50E+03** |
|  | Std | 4.47E+04 | 4.23E+03 | 3.82E+03 | **3.80E+02** | 2.97E+03 | 1.27E+03 | 9.13E+02 | 1.37E+04 |
| F15 | Mean | 7.58E+02 | 1.22E+02 | **1.00E+02** | 1.00E+02 | 1.00E+02 | 1.00E+02 | 1.00E+02 | 1.00E+02 |
|  | Std | 3.01E+03 | 2.77E+01 | **0.00E+00** | 1.88E-13 | 3.50E-10 | 1.59E-05 | 2.12E-10 | 4.85E-10 |

As shown in Table 4, when comparing AMPSO with MPSO and its comparison suite for the 10-*D* case, we can see that the mean values of AMPSO achieves solutions that are closer to the true optimum on 7 functions ($f_1, f_2, f_6, f_7, f_9, f_{14}$ and $f_{15}$), and the accuracy of the solution provided by AMPSO is at least an order of magnitude better than that of other PSO variants on $f_1, f_2$ and $f_{14}$. However, compared with MPSO on $f_3$, $f_4$ and $f_5$, DMSPSO on $f_8$ and CLPSO on $f_{11}$, our AMPSO is slightly weaker.

As shown in Table 5, for the 30-*D* case, compared to the counterparts, AMPSO obtains the best mean values in 10 out of 12 benchmark functions ($f_1, f_2, f_3, f_5, f_6, f_7, f_8, f_{11}, f_{14}$, and $f_{15}$). However, it is slightly weaker than MPSO on $f_4$ and $f_9$.

In conclusion, although AMPSO is weak in solving a few test functions, it has obvious advantages over MPSO and its comparison suite of PSO variants on the same 12 test functions of CEC2015 provided in [72]. Particularly, AMPSO has better solutions for high-dimensional complex functions.

Table 6, Table 7, Table 8 and Table 9 summarize the statistical values (*Best*, *Worst*, *Mean* and *Median*) of 8 algorithms on 15 test functions under *D* = 10 and *D* = 30

respectively when comparing AMPSO with PSODLS and its comparison suite, and the best results (the minimum value of each row) are shown in bold.

As presented in Table 6 and Table 7, for the 10-*D* case, the *Mean* value of AMPSO on the 6 functions $f_6, f_8, f_9, f_{10}, f_{13}, f_{14}$ and $f_{15}$ are the best among the 9 algorithms. Note that all these functions are hybrid or composition functions, and they are the most complex functions in the CEC 2015 test suite. In the other hand, PSODLS achieves the better outcomes on $f_4, f_7$ and $f_{11}$, DMSPSO finds better results on $f_3, f_5$, and $f_{12}$, while TLBO locates better local optima on $f_1$ and $f_2$.

For the 30-D case in Table 8 and Table 9, AMPSO achieved the best results on 11 functions $f_1, f_3, f_5, f_6, f_8, f_9, f_{10}, f_{12}, f_{13}, f_{14}$ and $f_{15}$ out of 15 benchmark functions. Except $f_1, f_3$ and $f_5$, all functions are hybrid or composition functions. This again proves the strength of AMPSO in solving complex optimization problems. However, AMPSO is inferior on $f_4, f_7$ and $f_{11}$ compared with PSODLS and inferior on $f_2$ with TLBO. In summary, the advantages of AMPSO in high-dimensional situations are still obvious, and 11 optimal solutions are obtained out of 15 functions.

In Table 10 and Table 11, statistical results in terms of mean, std, best, worst, and median are chosen as our evaluation criteria to conduct comparison of our algorithm with two recent PSO variants: EPSO and HCLPSO. From the tables, we can see that for the 10-*D* case, AMPSO obtains 6 best mean results on $f_3, f_6, f_9, f_{13}, f_{14}$ and $f_{15}$, equal to EPSO and HCLPSO on $f_9$ and $f_{15}$ (EPSO achieves the optimal standard deviation values on these two functions), while weaker than EPSO on $f_1, f_5, f_8, f_{10}$ and $f_{12}$ and HCLPSO on $f_2, f_4, f_7, f_{11}$. Similarly, for 30-D case, AMPSO achieves the optimal value on 8 functions ($f_1, f_3, f_6, f_8, f_{10}, f_{11}, f_{14}$ and $f_{15}$), while EPSO performs best on 4 functions ($f_9, f_{12}, f_{13}$ and $f_{15}$) and HCLPSO on 7 functions ($f_2, f_4, f_5, f_7, f_9, f_{12}$ and $f_{15}$). Although the optimal values AMPSO achieved in the 10-*D* condition is one less than that of EPSO and HCLPSO, in the 30-*D* AMPSO has four more functions better than EPSO and one more function better than HCLPSO.

**Table 6**

Comparison of AMPSO with PSODLS and its comparison suite for 10-D case on $f_1$-$f_8$

| Func | Item | GPSO* | MSPSO | FIPS | DTT-PSO | DMS-PSO | ABC | TLBO | PSODLS | AMPSO |
|---|---|---|---|---|---|---|---|---|---|---|
| F1 | Best | 8.37E-10 | 2.81E+02 | 4.26E+02 | 2.72E-07 | 2.11E-05 | 6.23E+03 | **0.00E+00** | 1.03E-11 | 1.93E+01 |
|    | Worst | 3.22E-04 | 6.36E+04 | 9.83E+04 | 1.12E+00 | 2.09E-02 | 2.21E+05 | **5.68E-14** | 2.31E-07 | 1.87E+04 |
|    | Average | 1.41E-05 | 1.97E+04 | 2.89E+04 | 5.99E-02 | 4.46E-03 | 8.58E+04 | **2.37E-14** | 4.39E-08 | 3.59E+03 |
|    | Median | 3.36E-07 | 1.62E+04 | 2.35E+04 | 7.20E-03 | 2.19E-03 | 9.16E+04 | **1.42E-14** | 2.24E-08 | 1.47E+03 |
| F2 | Best | 1.01E+00 | 5.50E+00 | 4.94E+01 | 1.74E+00 | 1.57E+01 | 1.05E+00 | 6.50E-01 | 5.00E-02 | **5.97E-03** |
|    | Worst | 1.32E+04 | 5.42E+03 | 1.53E+04 | 1.36E+04 | 1.42E+04 | 8.32E+02 | **3.00E+02** | 1.80E+03 | 1.20E+03 |
|    | Average | 3.35E+03 | 9.74E+02 | 3.66E+03 | 3.89E+03 | 4.43E+03 | 1.33E+02 | **9.41E+01** | 4.79E+02 | 2.76E+02 |
|    | Median | 1.17E+03 | 4.42E+02 | 2.51E+03 | 3.88E+03 | 2.33E+03 | 7.20E+01 | **5.12E+01** | 3.10E+02 | 8.85E+01 |
| F3 | Best | 2.00E+01 | 2.00E+01 | 1.66E+01 | 0.00E+00 | 0.00E+00 | 3.88E+00 | 2.32E+00 | 5.68E-14 | 2.81E+00 |
|    | Worst | 2.04E+01 | 2.03E+01 | 2.05E+01 | 2.05E+01 | 2.01E+01 | 2.01E+01 | 2.04E+01 | 2.03E+01 | **2.00E+01** |
|    | Average | 2.02E+01 | 2.02E+01 | 2.00E+01 | 1.97E+01 | **1.43E+01** | 1.76E+01 | 1.90E+01 | 1.79E+01 | 1.94E+01 |
|    | Median | 2.02E+01 | 2.02E+01 | 2.03E+01 | 2.03E+01 | **2.00E+01** | 2.01E+01 | 2.03E+01 | 2.02E+01 | **2.00E+01** |
| F4 | Best | 2.98E+00 | 2.01E+00 | 3.91E+00 | 1.35E+01 | 9.90E-01 | 4.03E+00 | 3.98E+00 | **3.00E-02** | 4.97E+00 |
|    | Worst | 1.19E+01 | 1.20E+01 | 1.70E+01 | 3.09E+01 | 6.92E+00 | 1.31E+01 | 1.68E+01 | **4.97E+00** | 2.39E+01 |
|    | Average | 7.03E+00 | 6.54E+00 | 1.15E+01 | 2.30E+01 | 3.96E+00 | 8.47E+00 | 9.47E+00 | **3.39E+00** | 1.41E+01 |
|    | Median | 6.96E+00 | 5.97E+00 | 1.25E+01 | 2.31E+01 | 3.98E+00 | 8.47E+00 | 9.52E+00 | **3.98E+00** | 1.34E+01 |
| F5 | Best | 1.02E+01 | 1.43E+01 | 2.25E+02 | 1.86E+01 | **2.50E-01** | 3.31E+01 | 1.54E+01 | 1.02E+01 | 6.83E+00 |
|    | Worst | 7.16E+02 | 5.07E+02 | 1.12E+03 | 1.29E+03 | **2.43E+02** | 3.48E+02 | 6.03E+02 | 2.53E+02 | 5.78E+02 |
|    | Average | 2.53E+02 | 2.77E+02 | 6.93E+02 | 8.95E+02 | **5.19E+01** | 1.91E+02 | 2.59E+02 | 1.19E+02 | 2.62E+02 |
|    | Median | 2.55E+02 | 2.76E+02 | 6.67E+02 | 8.74E+02 | **1.87E+01** | 1.89E+02 | 2.34E+02 | 1.29E+02 | 2.39E+02 |
| F6 | Best | 1.24E+02 | 7.60E+01 | 3.15E+02 | 1.97E+02 | 5.62E+01 | 4.41E+03 | 2.50E+02 | 3.58E+01 | **2.24E+00** |
|    | Worst | 2.20E+03 | 4.51E+03 | 2.29E+03 | 1.50E+03 | 1.09E+03 | 2.74E+05 | 1.31E+03 | 1.12E+03 | **7.32E+02** |
|    | Average | 6.64E+02 | 1.19E+03 | 8.11E+02 | 8.17E+02 | 3.73E+02 | 6.68E+04 | 6.73E+02 | 2.62E+02 | **2.37E+02** |
|    | Median | 5.62E+02 | 9.19E+02 | 6.96E+02 | 8.58E+02 | 2.77E+02 | 5.39E+04 | 5.90E+02 | **1.66E+02** | 1.81E+02 |
| F7 | Best | 3.90E-02 | 3.61E-01 | 1.06E+00 | 8.70E-02 | 8.40E-02 | 4.66E-01 | 1.28E-01 | 3.70E-02 | **3.67E-02** |
|    | Worst | 3.06E+00 | 2.54E+00 | 1.81E+00 | 3.04E+00 | 2.47E+00 | 1.62E+00 | 2.57E+00 | **1.09E+00** | 1.11E+00 |
|    | Average | 1.61E+00 | 1.43E+00 | 1.47E+00 | 1.32E+00 | 1.04E+00 | 9.50E-01 | 1.20E+00 | **4.60E-01** | 6.22E-01 |
|    | Median | 1.57E+00 | 1.54E+00 | 1.54E+00 | 1.24E+00 | 1.06E+00 | 9.20E-01 | 1.34E+00 | **1.40E-01** | 5.68E-01 |
| F8 | Best | 4.92E+01 | 1.65E+02 | 2.74E+02 | 1.51E+02 | 1.04E+02 | 6.36E+02 | 6.87E+01 | 4.13E+01 | **1.79E+01** |
|    | Worst | 6.14E+03 | 2.49E+03 | 3.74E+03 | 2.03E+03 | 4.90E+03 | 5.50E+04 | 8.98E+02 | **5.26E+02** | 1.68E+03 |
|    | Average | 2.41E+03 | 8.34E+02 | 1.12E+03 | 9.00E+02 | 1.89E+03 | 1.17E+04 | 2.99E+02 | 2.76E+02 | **2.39E+02** |
|    | Median | 1.91E+03 | 7.19E+02 | 9.09E+02 | 8.74E+02 | 1.70E+03 | 6.87E+03 | 1.80E+02 | 2.61E+02 | **1.45E+02** |

**Table 7**

Comparison of AMPSO with PSODLS and its comparison suite for 10-$D$ case on $f_9$-$f_{15}$

| Func | Item | GPSO* | MSPSO | FIPS | DTT-PSO | DMS-PSO | ABC | TLBO | PSODLS | AMPSO |
|---|---|---|---|---|---|---|---|---|---|---|
| F9 | Best | 1.02E+02 | 1.06E+02 | 1.02E+02 | 1.19E+02 | 1.00E+02 | **1.81E+01** | 1.05E+02 | 1.03E+02 | 1.00E+02 |
| | Worst | 2.01E+02 | 1.18E+02 | 1.30E+02 | 1.33E+02 | 1.10E+02 | 1.13E+02 | 1.12E+02 | 1.08E+02 | **1.00E+02** |
| | Average | 1.25E+02 | 1.10E+02 | 1.14E+02 | 1.28E+02 | 1.05E+02 | 1.01E+02 | 1.08E+02 | 1.07E+02 | **1.00E+02** |
| | Median | 1.12E+02 | 1.10E+02 | 1.12E+02 | 1.25E+02 | 1.05E+02 | 1.09E+02 | 1.08E+02 | 1.07E+02 | **1.00E+02** |
| F10 | Best | 7.96E+01 | 1.50E+02 | 1.15E+03 | 3.97E+02 | 4.88E+02 | 4.04E+02 | 2.04E+02 | **6.68E+01** | 2.23E+02 |
| | Worst | 2.46E+03 | 2.16E+03 | 2.18E+03 | 1.93E+03 | 2.11E+03 | 6.95E+03 | 1.19E+03 | 9.59E+02 | **7.63E+02** |
| | Average | 1.12E+03 | 9.52E+02 | 1.72E+03 | 1.29E+03 | 1.30E+03 | 2.70E+03 | 7.73E+02 | 4.24E+02 | **3.94E+02** |
| | Median | 1.19E+03 | 8.46E+02 | 1.73E+03 | 1.19E+03 | 1.29E+03 | 2.37E+03 | 8.77E+02 | **3.67E+02** | 3.79E+02 |
| F11 | Best | 5.80E-01 | 1.28E+00 | 3.53E+00 | 6.60E-01 | **2.90E-01** | 1.34E+00 | 8.80E-01 | 4.50E-01 | 6.03E+00 |
| | Worst | 3.68E+02 | 2.04E+02 | 2.09E+02 | 3.00E+02 | 3.00E+02 | 4.35E+00 | 2.00E+02 | **2.18E+00** | 3.01E+02 |
| | Average | 1.85E+02 | 1.05E+02 | 1.06E+02 | 1.85E+02 | 6.97E+01 | 3.09E+00 | 1.21E+02 | **1.23E+00** | 1.26E+02 |
| | Median | 2.00E+02 | 2.00E+02 | 1.03E+02 | 2.00E+02 | 6.59E+00 | 3.21E+00 | 2.00E+02 | **1.23E+00** | 1.20E+01 |
| F12 | Best | 1.02E+02 | 1.03E+02 | 1.02E+02 | 1.01E+02 | **1.00E+02** | 1.02E+02 | 1.01E+02 | 1.01E+02 | 1.02E+02 |
| | Worst | 1.06E+02 | 1.05E+02 | 1.03E+02 | 1.03E+02 | **1.02E+02** | 1.07E+02 | 1.06E+02 | 1.02E+02 | 1.05E+02 |
| | Average | 1.04E+02 | 1.04E+02 | 1.02E+02 | 1.02E+02 | **1.01E+02** | 1.06E+02 | 1.03E+02 | 1.02E+02 | 1.03E+02 |
| | Median | 1.03E+02 | 1.04E+02 | 1.02E+02 | 1.02E+02 | **1.01E+02** | 1.06E+02 | 1.02E+02 | 1.02E+02 | 1.03E+02 |
| F13 | Best | 1.13E+01 | 1.86E+01 | 1.92E+01 | 1.32E+01 | 7.35E+00 | 1.88E+01 | 1.79E+01 | 5.77E+00 | **2.82E-02** |
| | Worst | 3.70E+01 | 3.27E+01 | 3.26E+01 | 3.29E+01 | 3.18E+01 | 2.98E+01 | 2.91E+01 | 2.06E+01 | **3.06E-02** |
| | Average | 2.67E+01 | 2.74E+01 | 2.64E+01 | 2.40E+01 | 1.64E+01 | 2.47E+01 | 2.41E+01 | 1.58E+01 | **3.02E-02** |
| | Median | 2.71E+01 | 2.79E+01 | 2.69E+01 | 2.03E+01 | 1.56E+01 | 2.52E+01 | 2.51E+01 | 1.63E+01 | **3.05E-02** |
| F14 | Best | 1.00E+02 | 1.00E+02 | 1.00E+02 | 1.00E+02 | 1.00E+02 | 1.00E+02 | 1.00E+02 | 1.00E+02 | **3.12E-01** |
| | Worst | 2.73E+03 | 1.17E+02 | 2.02E+02 | 2.67E+03 | 2.74E+03 | **1.00E+02** | 2.73E+03 | **1.00E+02** | **1.00E+02** |
| | Average | 7.85E+02 | 1.03E+02 | 1.14E+02 | 2.72E+02 | 4.67E+02 | 1.00E+02 | 8.18E+02 | 1.00E+02 | **9.34E+01** |
| | Median | **1.00E+02** | 1.01E+02 | 1.01E+02 | **1.00E+02** | **1.00E+02** | **1.00E+02** | **1.00E+02** | **1.00E+02** | **1.00E+02** |
| F15 | Best | 2.05E+02 | 2.05E+02 | 2.05E+02 | 2.05E+02 | 2.05E+02 | 2.05E+02 | 2.05E+02 | 2.05E+02 | **1.00E+02** |
| | Worst | 2.05E+02 | 2.05E+02 | 2.05E+02 | 2.05E+02 | 2.05E+02 | 2.05E+02 | 2.56E+02 | 2.05E+02 | **1.00E+02** |
| | Average | 2.05E+02 | 2.05E+02 | 2.05E+02 | 2.05E+02 | 2.05E+02 | 2.05E+02 | 2.09E+02 | 2.05E+02 | **1.00E+02** |
| | Median | 2.05E+02 | 2.05E+02 | 2.05E+02 | 2.05E+02 | 2.05E+02 | 2.05E+02 | 2.05E+02 | 2.05E+02 | **1.00E+02** |

**Table 8**

Comparison of AMPSO with PSODLS and its comparison suite for 30-$D$ case on $f_1$-$f_8$

| Func | Item | GPSO* | MSPSO | FIPS | DTT-PSO | DMS-PSO | ABC | TLBO | PSODLS | AMPSO |
|---|---|---|---|---|---|---|---|---|---|---|
| F1 | Best | 2.21E+04 | 4.15E+06 | 2.57E+06 | 1.26E+05 | 1.58E+04 | 1.61E+06 | **1.52E+03** | 1.91E+04 | 1.25E+04 |
| | Worst | 2.03E+07 | 2.06E+07 | 1.11E+07 | 9.51E+06 | 1.16E+07 | 4.03E+06 | 5.13E+05 | 6.39E+05 | **9.09E+04** |
| | Average | 6.48E+06 | 1.03E+07 | 5.83E+06 | 1.28E+06 | 3.15E+06 | 2.92E+06 | 8.51E+04 | 1.82E+05 | **4.41E+04** |
| | Median | 5.81E+06 | 1.05E+07 | 5.75E+06 | 8.71E+05 | 2.69E+06 | 3.02E+06 | **3.46E+04** | 7.82E+04 | 4.09E+04 |
| F2 | Best | 6.48E-08 | 2.91E+06 | 2.06E+01 | 7.71E-03 | 5.77E-03 | 1.30E-01 | **1.71E-13** | 8.76E-11 | 5.42E-02 |
| | Worst | 1.43E-01 | 3.00E+07 | 1.16E+04 | 1.81E+04 | 1.24E+03 | 2.09E+02 | **1.80E-09** | 4.50E-06 | 7.45E+03 |
| | Average | 6.98E-03 | 1.26E+07 | 4.24E+03 | 5.20E+03 | 1.67E+02 | 4.02E+01 | **1.22E-10** | 1.27E-06 | 6.60E+02 |
| | Median | 1.51E-04 | 1.17E+07 | 3.02E+03 | 1.57E+03 | 3.76E+01 | 1.99E+01 | **3.69E-13** | 8.34E-07 | 2.57E+02 |
| F3 | Best | 2.04E+01 | 2.06E+01 | 2.09E+01 | 2.08E+01 | 2.02E+01 | 2.03E+01 | 2.08E+01 | 2.07E+01 | **2.00E+01** |
| | Worst | 2.10E+01 | 2.09E+01 | 2.10E+01 | 2.10E+01 | 2.05E+01 | 2.04E+01 | 2.10E+01 | 2.08E+01 | **2.00E+01** |
| | Average | 2.07E+01 | 2.08E+01 | 2.10E+01 | 2.10E+01 | 2.03E+01 | 2.04E+01 | 2.09E+01 | 2.08E+01 | **2.00E+01** |
| | Median | 2.08E+01 | 2.08E+01 | 2.10E+01 | 2.10E+01 | 2.03E+01 | 2.03E+01 | 2.09E+01 | 2.08E+01 | **2.00E+01** |
| F4 | Best | 4.18E+01 | 5.06E+01 | 7.13E+01 | 1.44E+02 | 1.79E+01 | 6.10E+01 | 6.37E+01 | **1.45E+01** | 1.04E+02 |
| | Worst | 1.10E+02 | 1.11E+02 | 1.47E+02 | 1.94E+02 | 6.17E+01 | 1.06E+02 | 1.43E+02 | **4.78E+01** | 1.84E+02 |
| | Average | 6.43E+01 | 8.78E+01 | 1.13E+02 | 1.72E+02 | 4.52E+01 | 8.42E+01 | 9.73E+01 | **3.96E+01** | 1.47E+02 |
| | Median | 5.97E+01 | 8.81E+01 | 1.14E+02 | 1.77E+02 | 4.74E+01 | 8.44E+01 | 9.65E+01 | **4.28E+01** | 1.49E+02 |
| F5 | Best | 1.42E+03 | 2.57E+03 | 5.19E+03 | 5.87E+03 | 2.55E+03 | 2.18E+03 | **1.13E+03** | 1.56E+03 | 1.58E+03 |
| | Worst | 4.58E+03 | 4.13E+03 | 7.02E+03 | 6.87E+03 | 3.99E+03 | 2.83E+03 | 6.73E+03 | **2.79E+03** | 2.92E+03 |
| | Average | 2.95E+03 | 3.43E+03 | 6.29E+03 | 6.34E+03 | 3.25E+03 | 2.48E+03 | 3.88E+03 | 2.42E+03 | **2.31E+03** |
| | Median | 3.02E+03 | 3.44E+03 | 6.35E+03 | 6.47E+03 | 3.22E+03 | 2.48E+03 | 3.17E+03 | 2.48E+03 | **2.31E+03** |
| F6 | Best | 9.62+e04 | 1.06E+05 | 1.03E+05 | 3.73E+04 | 1.44E+04 | 8.15E+05 | 2.57E+04 | 8.75E+03 | **4.38E+03** |
| | Worst | 1.46E+06 | 3.56E+06 | 1.05E+06 | 1.67E+06 | 1.72E+06 | 5.73E+06 | 1.09E+05 | **6.00E+04** | 9.05E+04 |
| | Average | 4.71E+05 | 8.52E+05 | 4.12E+05 | 3.68E+05 | 2.63E+05 | 3.25E+06 | 5.99E+04 | 3.71E+04 | **2.20E+04** |
| | Median | 2.44E+05 | 8.39E+05 | 3.84E+05 | 2.42E+05 | 1.88E+05 | 3.26E+06 | 5.76E+04 | 3.78E+04 | **1.60E+04** |
| F7 | Best | 5.18E+00 | 6.77E+00 | 3.80E+00 | 3.41E+00 | **2.76E+00** | 5.33E+00 | 5.88E+00 | 3.66E+00 | 6.07E+00 |
| | Worst | 1.74E+01 | 1.43E+01 | 1.11E+01 | 1.14E+01 | 7.83E+00 | 9.34E+00 | 1.41E+01 | **6.23E+00** | 1.00E+01 |
| | Average | 1.00E+01 | 1.04E+01 | 6.90E+00 | 6.56E+00 | 5.28E+00 | 7.78E+00 | 1.02E+01 | **5.09E+00** | 7.76E+00 |
| | Median | 9.29E+00 | 1.01E+01 | 6.98E+00 | 7.04E+00 | 5.29E+00 | 7.87E+00 | 1.05E+01 | **4.99E+00** | 7.64E+00 |
| F8 | Best | 3.40E+03 | 7.21E+03 | 1.07E+04 | 7.69E+03 | 3.04E+03 | 2.79E+05 | 2.88E+03 | **2.76E+03** | 5.04E+03 |
| | Worst | 4.14E+05 | 2.54E+05 | 2.73E+05 | 3.42E+05 | 1.09E+05 | 9.01E+05 | 7.60E+04 | 6.24E+04 | **2.99E+04** |
| | Average | 4.38E+04 | 8.28E+04 | 7.42E+04 | 7.71E+04 | 3.61E+04 | 5.41E+05 | 2.14E+04 | 1.76E+04 | **1.32E+04** |
| | Median | 3.10E+04 | 6.97E+04 | 6.16E+04 | 5.84E+04 | 3.07E+04 | 5.18E+05 | 1.72E+04 | 1.22E+04 | **1.20E+04** |

**Table 9**

Comparison of AMPSO with PSODLS and its comparison suite for 30-*D* case on $f_9$-$f_{15}$

| Func | Item | GPSO* | MSPSO | FIPS | DTTPSO | DMSPSO | ABC | TLBO | PSODLS | AMPSO |
|---|---|---|---|---|---|---|---|---|---|---|
| F9 | Best | 1.53E+02 | 1.69E+02 | 1.66E+02 | 2.01E+02 | 2.01E+02 | 1.69E+02 | 1.84E+02 | 1.58E+02 | **1.03E+02** |
| | Worst | 2.13E+02 | 2.02E+02 | 2.39E+02 | 2.94E+02 | 2.01E+02 | 2.02E+02 | 2.22E+02 | 2.01E+02 | **1.05E+02** |
| | Average | 1.94E+02 | 1.99E+02 | 2.01E+02 | 2.38E+02 | 2.01E+02 | 1.88E+02 | 2.02E+02 | 1.99E+02 | **1.04E+02** |
| | Median | 2.01E+02 | 2.02E+02 | 2.01E+02 | 2.01E+02 | 2.01E+02 | 1.87E+02 | 2.01E+02 | 2.01E+02 | **1.04E+02** |
| F10 | Best | 6.64E+03 | 2.84E+04 | 1.67E+04 | 3.42E+03 | 2.78E+03 | 6.61E+04 | 3.03E+03 | 2.36E+03 | **2.00E+03** |
| | Worst | 3.67E+05 | 8.80E+05 | 2.53E+05 | 2.16E+05 | 1.52E+05 | 1.05E+06 | 4.40E+04 | 4.48E+04 | **2.84E+04** |
| | Average | 8.77E+04 | 3.81E+05 | 1.06E+05 | 4.66E+04 | 3.58E+04 | 4.62E+05 | 1.63E+04 | 1.36E+04 | **1.24E+04** |
| | Median | 2.58E+04 | 3.56E+05 | 9.83E+04 | 1.62E+04 | 2.76E+04 | 4.37E+05 | 1.25E+04 | 1.25E+04 | **8.80E+03** |
| F11 | Best | 2.08E+02 | 2.12E+02 | 2.13E+02 | 2.03E+02 | 2.04E+02 | 2.08E+02 | 2.05E+02 | **2.03E+02** | 3.01E+02 |
| | Worst | 8.91E+02 | 7.70E+02 | 3.90E+02 | 4.73E+02 | 5.21E+02 | 2.26E+02 | 9.63E+02 | **2.24E+02** | 3.06E+02 |
| | Average | 5.81E+02 | 3.07E+02 | 3.20E+02 | 3.52E+02 | 4.14E+02 | 2.13E+02 | 6.12E+02 | **2.11E+02** | 3.04E+02 |
| | Median | 6.25E+02 | 2.42E+02 | 3.49E+02 | 4.00E+02 | 4.24E+02 | 2.19E+02 | 8.25E+02 | **2.10E+02** | 3.04E+02 |
| F12 | Best | 1.11E+02 | 1.12E+02 | 1.09E+02 | 1.07E+02 | 1.06E+02 | 1.10E+02 | 1.11E+02 | 1.08E+02 | **1.05E+02** |
| | Worst | 1.17E+02 | 1.15E+02 | 1.11E+02 | 1.11E+02 | 1.10E+02 | 1.12E+02 | 1.18E+02 | 1.10E+02 | **1.08E+02** |
| | Average | 1.13E+02 | 1.13E+02 | 1.10E+02 | 1.10E+02 | 1.08E+02 | 1.11E+02 | 1.14E+02 | 1.09E+02 | **1.07E+02** |
| | Median | 1.14E+02 | 1.13E+02 | 1.10E+02 | 1.10E+02 | 1.08E+02 | 1.11E+02 | 1.14E+02 | 1.09E+02 | **1.07E+02** |
| F13 | Best | 9.35E+01 | 9.59E+01 | 1.00E+02 | 1.00E+02 | 8.73E+01 | 9.78E+01 | 1.06E+02 | 8.27E+01 | **2.74E-02** |
| | Worst | 1.21E+02 | 1.19E+02 | 1.14E+02 | 1.20E+02 | 1.05E+02 | 1.10E+02 | 1.20E+02 | 1.00E+02 | **4.15E-02** |
| | Average | 1.07E+02 | 1.11E+02 | 1.08E+02 | 1.13E+02 | 9.59E+01 | 1.05E+02 | 1.12E+02 | 9.44E+01 | **3.33E-02** |
| | Median | 1.06E+02 | 1.13E+02 | 1.08E+02 | 1.14E+02 | 9.56E+01 | 1.05E+02 | 1.11E+02 | 9.52E+01 | **3.30E-02** |
| F14 | Best | 2.85E+04 | 2.98E+04 | 2.80E+04 | 2.84E+04 | 2.80E+04 | 2.65E+03 | 2.80E+04 | 2.80E+04 | **1.00E+02** |
| | Worst | 3.73E+04 | 3.25E+04 | 3.13E+04 | 3.78E+04 | 3.57E+04 | **3.08E+04** | 5.70E+04 | 3.10E+04 | 3.37E+04 |
| | Average | 3.22E+04 | 3.11E+04 | 3.02E+04 | 3.12E+04 | 3.10E+04 | 2.67E+04 | 3.46E+04 | 2.84E+04 | **7.50E+03** |
| | Median | 3.14E+04 | 3.11E+04 | 3.10E+04 | 3.04E+04 | 3.10E+04 | 2.74E+04 | 3.23E+04 | 2.84E+04 | **1.00E+02** |
| F15 | Best | 2.74E+02 | 2.78E+02 | 2.74E+02 | 2.73E+02 | 2.73E+02 | 2.74E+02 | 2.75E+02 | 2.73E+02 | **1.00E+02** |
| | Worst | 2.78E+02 | 2.84E+02 | 2.75E+02 | 2.74E+02 | 2.75E+02 | 2.75E+02 | 2.82E+02 | 2.74E+02 | **1.00E+02** |
| | Average | 2.75E+02 | 2.81E+02 | 2.74E+02 | 2.74E+02 | 2.74E+02 | 2.75E+02 | 2.78E+02 | 2.74E+02 | **1.00E+02** |
| | Median | 2.75E+02 | 2.81E+02 | 2.74E+02 | 2.73E+02 | 2.74E+02 | 2.75E+02 | 2.79E+02 | 2.74E+02 | **1.00E+02** |

**Table 10**

Comparison of AMPSO with EPSO and HCLPSO for 10-D and 30-*D* case on $f_1$-$f_8$

| Func | Item | 10-D EPSO | HCLPSO | AMSPO | Func | Item | 30-D EPSO | HCLPSO | AMPSO |
|---|---|---|---|---|---|---|---|---|---|
| F1 | Mean | **7.42E+02** | 3.28E+04 | 3.59E+03 | F1 | Mean | 5.88E+04 | 2.05E+05 | **4.41E+04** |
| | Std | **1.01E+03** | 2.42E+04 | 5.01E+03 | | Std | 4.43E+04 | 1.13E+05 | **2.08E+04** |
| | Best | **1.65E-01** | 4.79E+03 | 1.93E+01 | | Best | 1.76E+04 | 5.30E+04 | **1.25E+04** |
| | Worst | **5.09E+03** | 7.55E+04 | 1.87E+04 | | Worst | 2.47E+05 | 5.06E+05 | **9.09E+04** |
| | Median | **3.57E+02** | 2.60E+04 | 1.47E+03 | | Median | 5.51E+04 | 1.83E+05 | **4.09E+04** |
| F2 | Mean | 3.98E+02 | **1.41E+02** | 2.76E+02 | F2 | Mean | 2.91E+03 | **1.59E+02** | 6.60E+02 |
| | Std | 4.53E+02 | **1.43E+02** | 3.49E+02 | | Std | 3.00E+03 | **2.22E+02** | 1.45E+03 |
| | Best | 5.29E-02 | 1.32E+01 | **5.97E-03** | | Best | 4.79E+01 | 3.32E+00 | **5.42E-02** |
| | Worst | 2.04E+03 | **4.97E+02** | 1.20E+03 | | Worst | 1.20E+04 | **8.09E+02** | 7.45E+03 |
| | Median | 2.53E+02 | **7.34E+01** | 8.85E+01 | | Median | 1.99E+03 | **7.26E+01** | 2.57E+02 |
| F3 | Mean | 2.01E+01 | **1.94E+01** | **1.94E+01** | F3 | Mean | 2.02E+01 | 2.02E+01 | **2.00E+01** |
| | Std | **3.35E-02** | 3.67E+00 | 3.14E+00 | | Std | 5.18E-02 | 4.77E-02 | **8.68E-03** |
| | Best | 2.00E+01 | **2.27E-13** | 2.81E+00 | | Best | 2.01E+01 | 2.01E+01 | **2.00E+01** |
| | Worst | 2.02E+01 | 2.02E+01 | **2.00E+01** | | Worst | 2.03E+01 | 2.03E+01 | **2.00E+01** |
| | Median | 2.01E+01 | 2.01E+01 | **2.00E+01** | | Median | 2.02E+01 | 2.02E+01 | **2.00E+01** |
| F4 | Mean | 4.52E+00 | **4.04E+00** | 1.41E+01 | F4 | Mean | 4.66E+01 | **3.85E+01** | 1.47E+02 |
| | Std | **1.67E+00** | **1.67E+00** | 4.26E+00 | | Std | 1.12E+01 | **7.50E+00** | 2.24E+01 |
| | Best | **9.95E-01** | **9.95E-01** | 4.97E+00 | | Best | 2.29E+01 | **1.89E+01** | 1.04E+02 |
| | Worst | 8.97E+00 | **7.96E+00** | 2.39E+01 | | Worst | 6.77E+01 | **5.67E+01** | 1.84E+02 |
| | Median | 4.11E+00 | **3.98E+00** | 1.34E+01 | | Median | 4.78E+01 | **4.03E+01** | 1.49E+02 |
| F5 | Mean | **9.34E+01** | 1.16E+02 | 2.62E+02 | F5 | Mean | 1.87E+03 | **1.80E+03** | 2.31E+03 |
| | Std | **7.61E+01** | 8.57E+01 | 1.54E+02 | | Std | **2.75E+02** | 3.39E+02 | 3.55E+02 |
| | Best | 5.62E-01 | **3.12E-01** | 6.83E+00 | | Best | 1.30E+03 | **9.61E+02** | 1.58E+03 |
| | Worst | **2.98E+02** | 3.08E+02 | 5.78E+02 | | Worst | **2.30E+03** | 2.34E+03 | 2.92E+03 |
| | Median | **9.57E+01** | 1.30E+02 | 2.39E+02 | | Median | 1.92E+03 | **1.91E+03** | 2.31E+03 |
| F6 | Mean | 4.53E+02 | 4.17E+02 | **2.37E+02** | F6 | Mean | 2.39E+04 | 5.13E+04 | **2.20E+04** |
| | Std | 4.54E+02 | 3.31E+02 | **1.88E+02** | | Std | **1.78E+04** | 3.03E+04 | 2.25E+04 |
| | Best | 2.61E+01 | 2.61E+01 | **2.24E+00** | | Best | **2.71E+03** | 9.67E+03 | 4.38E+03 |
| | Worst | 1.61E+03 | 1.30E+03 | **7.32E+02** | | Worst | **6.77E+04** | 1.21E+05 | 9.05E+04 |
| | Median | 2.56E+02 | 3.33E+02 | **1.81E+02** | | Median | 1.77E+04 | 4.12E+04 | **1.60E+04** |
| F7 | Mean | 5.70E-01 | **3.79E-01** | 6.22E-01 | F7 | Mean | 5.97E+00 | **5.44E+00** | 7.76E+00 |
| | Std | 3.99E-01 | **3.49E-01** | 3.69E-01 | | Std | 1.43E+00 | 1.46E+00 | **9.65E-01** |
| | Best | 3.90E-02 | **2.68E-02** | 3.67E-02 | | Best | **3.15E+00** | 3.61E+00 | 6.07E+00 |
| | Worst | 1.12E+00 | **1.02E+00** | 1.11E+00 | | Worst | **8.34E+00** | 8.45E+00 | 1.00E+01 |
| | Median | 6.10E-01 | **2.39E-01** | 5.68E-01 | | Median | 6.03E+00 | **4.89E+00** | 7.64E+00 |
| F8 | Mean | **7.40E+01** | 1.22E+02 | 2.39E+02 | F8 | Mean | 1.53E+04 | 2.99E+04 | **1.32E+04** |
| | Std | **6.06E+01** | 1.26E+02 | 3.35E+02 | | Std | 9.25E+03 | 1.89E+04 | **6.14E+03** |
| | Best | **1.13E+00** | 7.97E+00 | 1.79E+01 | | Best | **2.05E+03** | 6.39E+03 | 5.04E+03 |
| | Worst | **2.19E+02** | 4.85E+02 | 1.68E+03 | | Worst | 3.48E+04 | 6.93E+04 | **2.99E+04** |
| | Median | **6.21E+01** | 7.10E+01 | 1.45E+02 | | Median | 1.33E+04 | 2.74E+04 | **1.20E+04** |

**Table 11**

Comparison of AMPSO with EPSO and HCLPSO for 10-*D* and 30-*D* case on $f_9$-$f_{15}$

| | | 10-D | | | | | 30-D | | |
|---|---|---|---|---|---|---|---|---|---|
| Func | Item | EPSO | HCLPSO | AMSPO | Func | Item | EPSO | HCLPSO | AMPSO |
| F9 | Mean | **1.00E+02** | **1.00E+02** | **1.00E+02** | F9 | Mean | **1.03E+02** | **1.03E+02** | 1.04E+02 |
| | Std | **2.98E-02** | 3.69E-02 | 8.40E-02 | | Std | 2.27E-01 | **1.97E-01** | 3.82E-01 |
| | Best | **1.00E+02** | **1.00E+02** | **1.00E+02** | | Best | **1.02E+02** | **1.02E+02** | 1.03E+02 |
| | Worst | **1.00E+02** | **1.00E+02** | **1.00E+02** | | Worst | **1.03E+02** | **1.03E+02** | 1.05E+02 |
| | Median | **1.00E+02** | **1.00E+02** | **1.00E+02** | | Median | **1.03E+02** | **1.03E+02** | 1.04E+02 |
| F10 | Mean | **2.99E+02** | 3.40E+02 | 3.94E+02 | F10 | Mean | 1.48E+04 | 2.71E+04 | **1.24E+04** |
| | Std | **5.80E+01** | 7.71E+01 | 1.29E+02 | | Std | 7.80E+03 | 1.32E+04 | **7.40E+03** |
| | Best | **2.19E+02** | 2.31E+02 | 2.23E+02 | | Best | 2.01E+03 | 5.18E+03 | **2.00E+03** |
| | Worst | **4.11E+02** | 4.90E+02 | 7.63E+02 | | Worst | 2.97E+04 | 5.72E+04 | **2.84E+04** |
| | Median | **2.84E+02** | 3.34E+02 | 3.79E+02 | | Median | 1.43E+04 | 2.86E+04 | **8.80E+03** |
| F11 | Mean | 1.21E+02 | **5.30E+01** | 1.26E+02 | F11 | Mean | 3.44E+02 | 3.06E+02 | **3.04E+02** |
| | Std | 1.49E+02 | **1.12E+02** | 1.45E+02 | | Std | **1.24E+02** | 3.90E+00 | 1.29E+00 |
| | Best | **6.11E-01** | 1.80E+00 | 6.03E+00 | | Best | **3.01E+02** | 3.02E+02 | **3.01E+02** |
| | Worst | **3.00E+02** | **3.00E+02** | 3.01E+02 | | Worst | 8.22E+02 | 3.22E+02 | **3.06E+02** |
| | Median | **2.19E+00** | 3.78E+00 | 1.20E+01 | | Median | **3.04E+02** | 3.05E+02 | **3.04E+02** |
| F12 | Mean | **1.01E+02** | 1.02E+02 | 1.03E+02 | F12 | Mean | **1.05E+02** | **1.05E+02** | 1.07E+02 |
| | Std | **3.27E-01** | 3.72E-01 | 8.50E-01 | | Std | 7.26E-01 | **5.99E-01** | 8.44E-01 |
| | Best | **1.01E+02** | **1.01E+02** | 1.02E+02 | | Best | **1.04E+02** | **1.04E+02** | 1.05E+02 |
| | Worst | **1.02E+02** | **1.02E+02** | 1.05E+02 | | Worst | **1.06E+02** | 1.07E+02 | 1.08E+02 |
| | Median | **1.01E+02** | 1.02E+02 | 1.03E+02 | | Median | **1.05E+02** | **1.05E+02** | 1.07E+02 |
| F13 | Mean | 3.04E-02 | 3.04E-02 | **3.02E-02** | F13 | Mean | **2.80E-02** | 2.81E-02 | 3.33E-02 |
| | Std | **7.85E-05** | 3.65E-04 | 7.42E-04 | | Std | 1.20E-03 | **1.17E-03** | 4.17E-03 |
| | Best | 3.03E-02 | 2.88E-02 | **2.82E-02** | | Best | 2.60E-02 | **2.59E-02** | 2.74E-02 |
| | Worst | **3.05E-02** | 3.06E-02 | 3.06E-02 | | Worst | 3.12E-02 | **3.10E-02** | 4.15E-02 |
| | Median | **3.04E-02** | 3.05E-02 | 3.05E-02 | | Median | **2.78E-02** | 2.83E-02 | 3.30E-02 |
| F14 | Mean | 6.97E+02 | 7.29E+02 | **9.34E+01** | F14 | Mean | 3.16E+04 | 3.15E+04 | **7.50E+03** |
| | Std | 1.49E+03 | 1.17E+03 | **2.53E+01** | | Std | 4.82E+02 | **2.69E+02** | 1.37E+04 |
| | Best | 1.00E+02 | 1.00E+02 | **3.12E-01** | | Best | 3.11E+04 | 3.12E+04 | **1.00E+02** |
| | Worst | 6.68E+03 | 2.96E+03 | **1.00E+02** | | Worst | 3.30E+04 | **3.23E+04** | 3.37E+04 |
| | Median | **1.00E+02** | **1.00E+02** | **1.00E+02** | | Median | 3.15E+04 | 3.14E+04 | **1.00E+02** |
| F15 | Mean | **1.00E+02** | **1.00E+02** | **1.00E+02** | F15 | Mean | **1.00E+02** | **1.00E+02** | **1.00E+02** |
| | Std | **5.77E-14** | 1.25E-13 | 1.79E-09 | | Std | **1.55E-13** | 3.26E-13 | 4.85E-10 |
| | Best | **1.00E+02** | **1.00E+02** | **1.00E+02** | | Best | **1.00E+02** | **1.00E+02** | **1.00E+02** |
| | Worst | **1.00E+02** | **1.00E+02** | **1.00E+02** | | Worst | **1.00E+02** | **1.00E+02** | **1.00E+02** |
| | Median | **1.00E+02** | **1.00E+02** | **1.00E+02** | | Median | **1.00E+02** | **1.00E+02** | **1.00E+02** |

The Matlab code of our proposed approach is available in [79].

## 5. Conclusion and Future Work

We proposed an artificial multi-swarm PSO (AMPSO) to improve the performance of PSO, the multi-swarm system consists of an exploration swarm with many equal-sized subswarms, an artificial exploitation swarm generated from the best particle of the exploration swarm, and an artificial convergence swarm generated from the best particle of the exploitation swarm. To guarantee the success of the strategy, the Exploration Swarm and the Exploitation Swarm are conducted alternatively, and a novel diversity method is also applied for the swarms; to compensate the drawback of slow convergence of a swarm with large diversity, two swarm update techniques (i.e., partial swarm reconstruction and swarm reconstruction) are applied to the Exploitation Swarm and the Convergence Swarm to guarantee that AMPSO can achieve nice results within a fixed number of iterations.

We validate our AMPSO by comparing it with a set of comprehensive 16 algorithms, including the most recently well-performing PSO variants and some other non-PSO optimization algorithms under the same criteria.

As presented above, for the 30-*D* case, AMPSO obtains 10 best average solutions out of 12 benchmark functions when compared with MPSO, 11 out of 15 benchmark functions when compared with PSODLS, and 8 out of 15 benchmark functions when compared with EPSO and HCLPSO. We conclude that AMPSO is excellent in solving complex high-dimensional problems.

However, there is still much room for improvement when our AMPSO is dealing with lower dimensions. For example, for the 10-*D* case, 7 (out of 12), 6 (out of 15) and 6 (out of 15) best mean values can be obtained by AMPSO in the comparison experiments with MPSO, PSODLS and EPSO and HCLPSO respectively. Although the results are nice when compared with other algorithms, but less significant when compared with the performance of AMPSO in the 30-*D* case. Also, AMPSO has a large *std* in Table 4 and Table 5, which begs for a future research to reduce it.

We have also tested our algorithm on CEC2013 (with simpler benchmark functions than CEC2015) in the case of dimension 30 and found that in terms of accuracy, it is slightly weaker than the state-of-the-art algorithms such as MPSO [72] and MSPSO [80]. Our algorithm still has some room of improvement on high-dimensional but simple functions and this is also worthy of future research.

Furthermore, although adjusting the inertia weight has been proved to be effective in balancing the capabilities of the exploration and exploitation, more parameters (e.g., $c_1$ and $c_2$) can be designed into adaptive settings for the swarms to achieve better results.

Finally, our AMPSO may be incorporated with other nice local search algorithms such as GA and SAA to improve its ability in producing best particles for the transmissions from one swarm to another.

**Acknowledgement:**
This work was partially funded by the Shanghai International Science and Technology Cooperation Fund No. 18510745700 and the Science and Technology Development Fund, Macau SAR (Grant no. 0018/2019/AKP, and 0008/2019/AGJ)

**Reference:**
[1] R. Eberhart, J. Kennedy, A new optimizer using particle swarm theory, in: MHS'95.

Proc. Sixth Int. Symp. Micro Mach. Hum. Sci., 1995: pp. 39–43.

[2] J. Kennedy, R.C. Eberhart, Particle swarm optimization, Proc. of IEEE Int. Conf. on Neural Network, CNN'95 (1995) 1942–1948.

[3] X. Xu, Y. Tang, J. Li, C. Hua, X. Guan, Dynamic multi-swarm particle swarm optimizer with cooperative learning strategy, Appl. Soft Comput. 29 (2015) 169–183.

[4] T. Blackwell, A study of collapse in bare bones particle swarm optimization, IEEE Trans. Evol. Comput. 16 (2012) 354–372.

[5] P. Kuila, P.K. Jana, Energy efficient clustering and routing algorithms for wireless sensor networks: Particle swarm optimization approach, Eng. Appl. Artif. Intell. 33 (2014) 127–140.

[6] M. Shen, Z.H. Zhan, W.N. Chen, Y.J. Gong, J. Zhang, Y. Li, Bi-velocity discrete particle swarm optimization and its application to multicast routing problem in communication networks, IEEE Trans. Ind. Electron. 61 (2014) 7141–7151.

[7] Y. Zhang, S. Wang, G. Ji, Z. Dong, An MR Brain Images Classifier System via Particle Swarm Optimization and Kernel Support Vector Machine, Sci. World J. 2013 (2013) 130-134.

[8] L.D. Dhinesh Babu, P. Venkata Krishna, Honey bee behavior inspired load balancing of tasks in cloud computing environments, Appl. Soft Comput. 13 (2013) 2292–2303.

[9] Z. J. Wang, Z. H. Zhan, S. Kwong, H. Jin, J. Zhang, Adaptive Granularity Learning Distributed Particle Swarm Optimization for Large-Scale Optimization, IEEE Trans. Cybern. (2020) 1–14.

[10] W. Deng, H. Zhao, X. Yang, J. Xiong, M. Sun, B. Li, Study on an improved adaptive PSO algorithm for solving multi-objective gate assignment, 59 (2017) 288–302.

[11] M. Sharafi, T.Y. ELMekkawy, Multi-objective optimal design of hybrid renewable energy systems using PSO-simulation based approach, Renew. Energy. 68 (2014) 67–79.

[12] F.J. Cabrerizo, E. Herrera-Viedma, W. Pedrycz, A method based on PSO and granular computing of linguistic information to solve group decision making problems defined in heterogeneous contexts, Eur. J. Oper. Res. 230 (2013) 624–633.

[13] X. Zhang, K. J. Du, Z. H. Zhan, S. Kwong, T. L. Gu, J. Zhang, Cooperative Coevolutionary Bare-Bones Particle Swarm Optimization With Function Independent Decomposition for Large-Scale Supply Chain Network Design With Uncertainties, IEEE Trans. Cybern. (2019) 1–15.

[14] Z.H. Zhan, J. Zhang, Y. Li, H.S.H. Chung, Adaptive particle swarm optimization, IEEE Trans. Syst. Man, Cybern. Part B Cybern. 39 (2009) 1362–1381.

[15] J. Gou, Y.X. Lei, W.P. Guo, C. Wang, Y.Q. Cai, W. Luo, A novel improved particle swarm optimization algorithm based on individual difference evolution, Appl. Soft Comput. 57 (2017) 468–481.

[16] B. Niu, Y. Zhu, X. He, H. Wu, MCPSO: A multi-swarm cooperative particle swarm optimizer, Appl. Math. Comput. 185 (2007) 1050–1062.

[17] L. Wang, B. Yang, Y. Chen, Improving particle swarm optimization using multi-layer searching strategy, Inf. Sci. 274 (2014) 70–94.

[18] F. Han, Q. Liu, A diversity-guided hybrid particle swarm optimization based on gradient search, Neurocomputing. 137 (2014) 234–240.

[19] F. Zhao, J. Tang, J. Wang, Jonrinaldi, An improved particle swarm optimization with decline disturbance index (DDPSO) for multi-objective job-shop scheduling problem, Comput. Oper. Res. 45 (2014) 38–50.

[20] Y. Chen, L. Li, H. Peng, J. Xiao, Q. Wu, Dynamic multi-swarm differential learning particle swarm optimizer, Swarm Evol. Comput. 39 (2018) 209–221.

[21] N. Lynn, M.Z. Ali, P.N. Suganthan, Population topologies for particle swarm optimization and differential evolution, Swarm Evol. Comput. 39 (2018) 24–35.

[22] L. Zhang, Y. Tang, C. Hua, X. Guan, A new particle swarm optimization algorithm with adaptive inertia weight based on Bayesian techniques, Appl. Soft Comput. 28 (2015) 138–149.

[23] M.R. Tanweer, S. Suresh, N. Sundararajan, Self regulating particle swarm optimization algorithm, Inf. Sci. 294 (2015) 182–202.

[24] M. Taherkhani, R. Safabakhsh, A novel stability-based adaptive inertia weight for particle swarm optimization, Appl. Soft Comput. 38 (2016) 281–295.

[25] J. Kennedy, R. Mendes, Population structure and particle swarm performance, in: Proceedings of the Congress on Evolutionary Computation, 2002, pp. 1671–1676.

[26] R. Mendes, J. Kennedy, J. Neves, The fully informed particle swarm: simpler, maybe better, IEEE Trans. Evol. Comput. 8 (2004) 204–210.

[27] Y. Cooren, M. Clerc, P. Siarry, Performance evaluation of TRIBES, an adaptive particle swarm optimization algorithm, Swarm Intell. 3 (2009) 149–178.

[28] J. Kennedy, Small worlds and mega-minds: Effects of neighborhood topology on particle swarm performance, in: Proc. 1999 Congr. Evol. Comput. CEC 1999, IEEE Computer Society, 1999: pp. 1931–1938.

[29] P.N. Suganthan, Particle swarm optimiser with neighbourhood operator, Proc. 1999 Congr. Evol. Comput. CEC 1999. 3 (1999) 1958–1962.

[30] M.R. Bonyadi, X. Li, Z. Michalewicz, A hybrid particle swarm with a time-adaptive topology for constrained optimization, Swarm Evol. Comput. 18 (2014) 22–37.

[31] S. Mirjalili, S.Z. Mohd Hashim, H. Moradian Sardroudi, Training feedforward neural networks using hybrid particle swarm optimization and gravitational search algorithm, Appl. Math. Comput. 218 (2012) 11125–11137.

[32] J.R. Zhang, J. Zhang, T.M. Lok, M.R. Lyu, A hybrid particle swarm optimization-back-propagation algorithm for feedforward neural network training, Appl. Math. Comput. 185 (2007) 1026–1037.

[33] A.A. Nagra, F. Han, Q.H. Ling, S. Mehta, An Improved Hybrid Method Combining Gravitational Search Algorithm With Dynamic Multi Swarm Particle Swarm Optimization, IEEE Access. 7 (2019) 50388–50399.

[34] Z.H. Zhan, J. Zhang, Y. Li, Y.H. Shi, Orthogonal learning particle swarm optimization, IEEE Trans. Evol. Comput. 15 (2011) 832–847.

[35] H. Garg, A hybrid PSO-GA algorithm for constrained optimization problems, Appl.

Math. Comput. 274 (2016) 292–305.

[36] L.L. Liu, R.S. Hu, X.P. Hu, G.P. Zhao, S. Wang, A hybrid PSO-GA algorithm for job shop scheduling in machine tool production, Int. J. Prod. Res. 53 (2015) 5755–5781.

[37] V. Plevris, M. Papadrakakis, A Hybrid Particle Swarm-Gradient Algorithm for Global Structural Optimization, Comput. Civ. Infrastruct. Eng. 26 (2011) 48–68.

[38] N. Singh, S.B. Singh, Hybrid Algorithm of Particle Swarm Optimization and Grey Wolf Optimizer for Improving Convergence Performance, J. Appl. Math. 2017 (2017).

[39] M. Raju, M.K. Gupta, N. Bhanot, V.S. Sharma, A hybrid PSO–BFO evolutionary algorithm for optimization of fused deposition modelling process parameters, J. Intell. Manuf. 30 (2019) 2743–2758.

[40] S.N. Sivanandam, P. Visalakshi, Dynamic task scheduling with load balancing using parallel orthogonal particle swarm optimization, Int. J. Bio-Inspired Comput. 1 (2009) 276–286.

[41] S. Sengupta, S. Basak, R. Peters, Particle Swarm Optimization: A Survey of Historical and Recent Developments with Hybridization Perspectives, Mach. Learn. Knowl. Extr. 1 (2018) 157–191.

[42] D. Wang, D. Tan, L. Liu, Particle swarm optimization algorithm: an overview, Soft Comput. 22 (2018) 387–408.

[43] Y. Shi, R.C. Eberhart, A modified particle swarm optimizer, 1998 IEEE Int. Conf. Evol. Comput. (1998) 69–73.

[44] M. Chih, C.J. Lin, M.S. Chern, T.Y. Ou, Particle swarm optimization with time-varying acceleration coefficients for the multidimensional knapsack problem, Appl. Math. Model. 38 (2014) 1338–1350.

[45] A. Ratnaweera, S.K. Halgamuge, H.C. Watson, Self-organizing hierarchical particle swarm optimizer with time-varying acceleration coefficients, IEEE Trans. Evol. Comput. 8 (2004) 240–255.

[46] M.A. Potter, K.A. De Jong, A cooperative coevolutionary approach to function optimization, International Conference on Parallel Problem Solving from Nature, Springer (1994), pp. 249-257

[47] F. van den Bergh, A.P. Engelbrecht, A cooperative approach to particle swarm optimization, IEEE Trans. Evol. Comput. 8 (2004) 225–239.

[48] T. Blackwell, J. Branke, Multi-swarm optimization in dynamic environments, Lect. Notes Comput. Sci. (Including Subser. Lect. Notes Artif. Intell. Lect. Notes Bioinformatics). 3005 (2004) 489–500.

[49] J.J. Liang, P.N. Suganthan, Dynamic multi-swarm particle swarm optimizer with local search, 2005 IEEE Congr. Evol. Comput. IEEE CEC 2005. Proc. 1 (2005) 522–528.

[50] S.Z. Zhao, J.J. Liang, P.N. Suganthan, M.F. Tasgetiren, Dynamic multi-swarm particle swarm optimizer with local search for large scale global optimization, 2008 IEEE Congr. Evol. Comput. CEC 2008. (2008) 3845–3852.

[51] J. Zhang, X. Ding, A multi-swarm self-adaptive and cooperative particle swarm optimization, Eng. Appl. Artif. Intell. 24 (2011) 958–967.

[52] S.Z. Zhao, P.N. Suganthan, Q.K. Pan, M. Fatih Tasgetiren, Dynamic multi-swarm particle swarm optimizer with harmony search, Expert Syst. Appl. 38 (2011) 3735–3742.

[53] W. Ye, W. Feng, S. Fan, A novel multi-swarm particle swarm optimization with dynamic learning strategy, Appl. Soft Comput. J. 61 (2017) 832–843.

[54] Z.J. Wang, Z.H. Zhan, W.J. Yu, Y. Lin, J. Zhang, T.L. Gu, J. Zhang, Dynamic Group Learning Distributed Particle Swarm Optimization for Large-Scale Optimization and Its Application in Cloud Workflow Scheduling, IEEE Trans. Cybern. 50 (2020) 2715–2729.

[55] Y. Wu, X.Z. Gao, X.L. Huang, K. Zenger, A hybrid optimization method of Particle Swarm Optimization and Cultural Algorithm, in: Proc. - 2010 6th Int. Conf. Nat. Comput. ICNC 2010, 2010: pp. 2515–2519.

[56] M. Xu, X. You, S. Liu, A Novel Heuristic Communication Heterogeneous Dual Population Ant Colony Optimization Algorithm, IEEE Access. 5 (2017) 18506–18515.

[57] N. Netjinda, T. Achalakul, B. Sirinaovakul, Particle Swarm Optimization inspired by starling flock behavior, Appl. Soft Comput. 35 (2015) 411–422.

[58] W. Fang, J. Sun, H. Chen, X. Wu, A decentralized quantum-inspired particle swarm optimization algorithm with cellular structured population, Inf. Sci. 330 (2016) 19–48.

[59] W.X. Zhang, W.N. Chen, J. Zhang, A dynamic competitive swarm optimizer based-on entropy for large scale optimization, Proc. 8th Int. Conf. Adv. Comput. Intell. ICACI 2016. (2016) 365–371.

[60] M. Ran, Q. Wang, C. Dong, A dynamic search space Particle Swarm Optimization algorithm based on population entropy, 26th Chinese Control Decis. Conf. CCDC 2014. (2014) 4292–4296.

[61] K. Tang, Z. Li, L. Luo, B. Liu, Multi-strategy adaptive particle swarm optimization for numerical optimization, Eng. Appl. Artif. Intell. 37 (2015) 9–19.

[62] J. Zhu, Y. Lin, W. Lei, Y. Liu, M. Tao, Optimal household appliances scheduling of multiple smart homes using an improved cooperative algorithm, Energy. 171 (2019) 944–955.

[63] J. Riget, J. Vesterstroem, A Diversity-guided Particle Swarm Optimizer--The ARPSO, Technical Report, Department of Computer Science, University of Aarhus (2002)

[64] O. Olorunda, A.P. Engelbrecht, Measuring exploration/exploitation in particle swarms using swarm diversity, 2008 IEEE Congr. Evol. Comput. CEC 2008. (2008) 1128–1134.

[65] C.E. Shannon, A mathematical theory of communication, Bell Syst. Tech. J. 27 (1948) 379–423.

[66] S. Cheng, Y. Shi, Diversity control in particle swarm optimization, in: 2011 IEEE Symp. Swarm Intell., 2011: pp. 1–9.

[67] W. Guo, L. Zhu, L. Wang, Q. Wu, F. Kong, An entropy-assisted particle swarm optimizer for large-scale optimization problem, Mathematics. 7 (2019) 1–12.

[68] Y. Cao, H. Zhang, W. Li, M. Zhou, Y. Zhang, W.A. Chaovalitwongse,

Comprehensive Learning Particle Swarm Optimization Algorithm With Local Search for Multimodal Functions, IEEE Trans. Evol. Comput. 23 (2019) 718–731.

[69] J.J. Liang, B. Qu, P. Suganthan, Problem definitions and evaluation criteria for the cec 2015 competition on learning-based real-parameter single objective optimization, Tech. rep., Nanyang Technological University (Singapore) and Zhengzhou University (China), Available at: www.ntu.edu.sg/home/ epnsugan/ (Nov. 2014).

[70] D. Bratton, J. Kennedy, Defining a standard for particle swarm optimization, Proceedings of the IEEE Swarm Intelligence Symposium (SIS'07) (2007), pp. 120-127.

[71] J.J. Liang, A.K. Qin, P.N. Suganthan, S. Baskar, Comprehensive learning particle swarm optimizer for global optimization of multimodal functions, IEEE Trans. Evol. Comput. 10 (2006) 281–295.

[72] D. Tian, Z. Shi, MPSO: Modified particle swarm optimization and its applications, Swarm Evol. Comput. 41 (2018) 49–68.

[73] W. Der Chang, A modified particle swarm optimization with multiple subpopulations for multimodal function optimization problems, Appl. Soft Comput. 33 (2015) 170–182.

[74] L. Wang, B. Yang, J. Orchard, Particle swarm optimization using dynamic tournament topology, Appl. Soft Comput. 48 (2016) 584–596.

[75] D. Karaboga, B. Basturk, On the performance of artificial bee colony (ABC) algorithm, Appl. Soft Comput. 8 (2008) 687–697.

[76] R.V. Rao, V.J. Savsani, D.P. Vakharia, Teaching-learning-based optimization: a novel method for constrained mechanical design optimization problems, Comput. Aided Des. 43 (2011) 303–315.

[77] N. Lynn, P.N. Suganthan, Ensemble particle swarm optimizer, Appl. Soft Comput. J. 55 (2017) 533–548.

[78] N. Lynn, P.N. Suganthan, Heterogeneous comprehensive learning particle swarm optimization with enhanced exploration and exploitation, Swarm Evol. Comput. 24 (2015) 11–24.

[79] Haohao Zhou, Zhi-Hui Zhan, Zhi-Xin Yang, Xiangzhi Wei, https://github.com/elmerzhouhaohao/AMPSO, 2020 (accessed 17 June 2020).

[80] X. Xia, L. Gui, Z.H. Zhan, A multi-swarm particle swarm optimization algorithm based on dynamical topology and purposeful detecting, Appl. Soft Comput. 67 (2018) 126–140.